\documentclass[journal]{IEEEtran}

\usepackage[utf8]{inputenc} 
\usepackage[T1]{fontenc}    
\usepackage{hyperref}       
\usepackage{url}            
\usepackage{booktabs}       
\usepackage{amsfonts}       
\usepackage{nicefrac}       
\usepackage{microtype}      
\usepackage{lipsum}
\usepackage{fancyhdr}       
\usepackage{graphicx}       
\usepackage{subcaption}
\graphicspath{{media/}}     

\usepackage{multirow}
\usepackage[table,xcdraw]{xcolor}
\usepackage{amsmath}
\usepackage[normalem]{ulem}
\useunder{\uline}{\ul}{}

\usepackage{tocloft}
\usepackage{etoc}

\usepackage{dblfloatfix}
\usepackage{siunitx}
\usepackage{cite}

\usepackage[italicComments=true]{algpseudocodex}
\usepackage{algorithm}

\begin{document}

\bstctlcite{IEEEexample:BSTcontrol}

\DeclareSIUnit{\points}{pts}

\title{TanDepth: Leveraging Global DEMs for \\ Metric Monocular Depth Estimation in UAVs}

\author{Horatiu~Florea
        and~Sergiu~Nedevschi
\thanks{The research reported in this paper was supported by Lockheed Martin. \newline The authors are with the Department of Computer Science, Technical
University of Cluj-Napoca, 26-28 Gh. Baritiu Street, 400114, Cluj-Napoca, Romania (Corresponding Author: Sergiu Nedevschi e-mail:
FirstName.LastName@cs.utcluj.ro)}
}

\maketitle

\begin{abstract}
Aerial scene understanding systems face stringent payload restrictions and must often rely on monocular depth estimation for modeling scene geometry, which is an inherently ill-posed problem. 
Moreover, obtaining accurate ground truth data required by learning-based methods raises significant additional challenges in the aerial domain. 
Self-supervised approaches can bypass this problem, at the cost of providing only up-to-scale results. 
Similarly, recent supervised solutions which make good progress towards zero-shot generalization also provide only relative depth values. 
This work presents TanDepth, a practical scale recovery method for obtaining metric depth results from relative estimations at inference-time, irrespective of the type of model generating them. 
Tailored for Unmanned Aerial Vehicle (UAV) applications, our method leverages sparse measurements from Global Digital Elevation Models (GDEM) by projecting them to the camera view using extrinsic and intrinsic information. An adaptation to the Cloth Simulation Filter is presented, which allows selecting ground points from the estimated depth map to then correlate with the projected reference points.
We evaluate and compare our method against alternate scaling methods adapted for UAVs, on a variety of real-world scenes. Considering the limited availability of data for this domain, we construct and release a comprehensive, depth-focused extension to the popular UAVid dataset to further research.
\end{abstract}


\IEEEpeerreviewmaketitle

\section{Introduction}
Recent years have brought constant technical progress and increasing economies of scale in the Unmanned Aerial Vehicle (UAV) market, accelerating research interest \cite{nex2022uav}. 
Their roles expand beyond industrial settings like surveying and agriculture to fields of crisis response, such as wildfires \cite{akhloufi2021appfire} or search and rescue \cite{lyu2023appsar}. 
To maximize the reliability and utility of UAVs, advanced perception systems encompassing tasks such as object detection, localization and mapping are being actively developed.
A key function of such systems is to generate a consistent, accurate representation of the 3D geometry in the flight area modelling terrain, natural and man-made structures, as well as actors present in the scene.

\begin{figure}
\footnotesize
    \centering
	\includegraphics[width=9 cm]{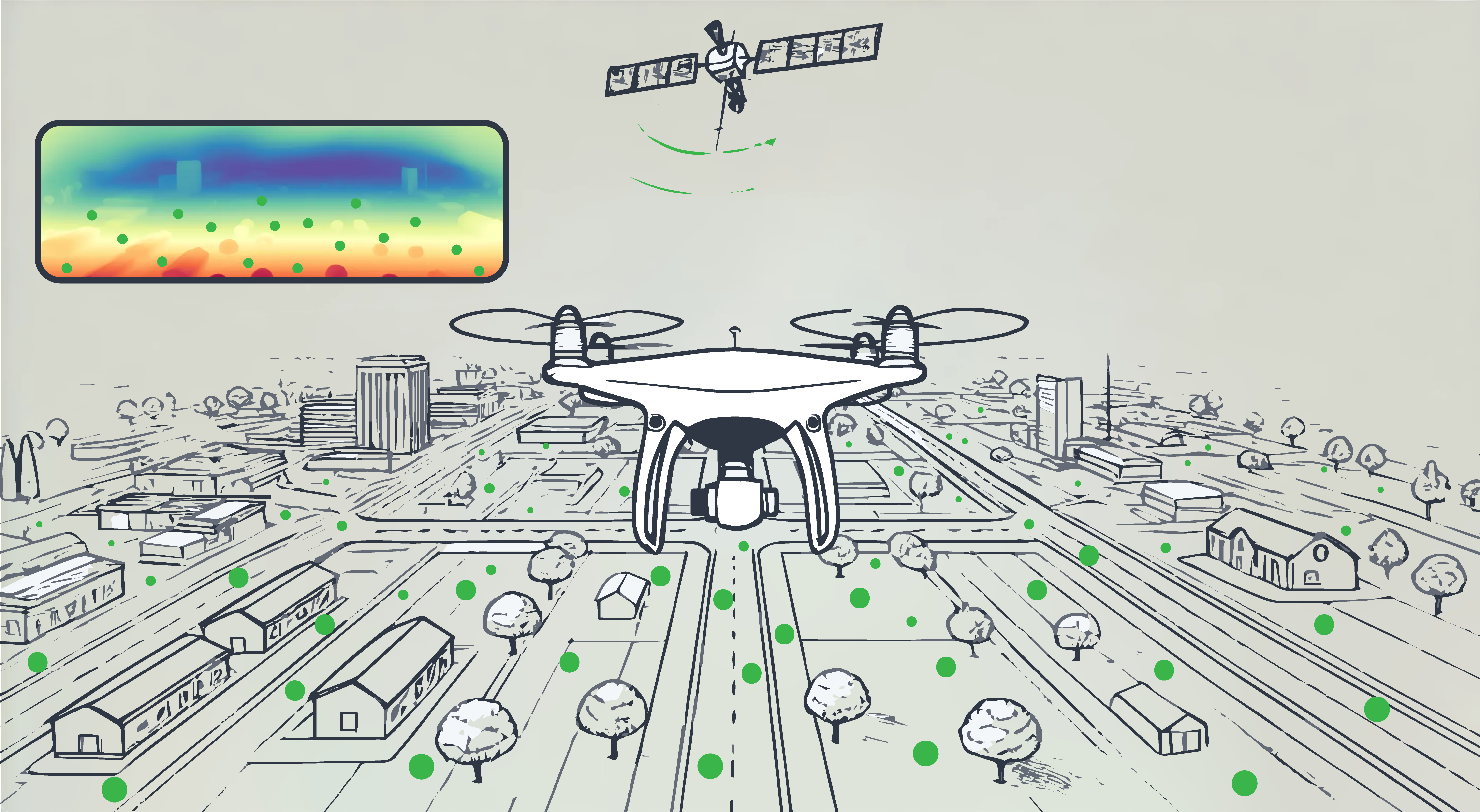}
        \caption{TanDepth enables relative depth maps (left window) used by UAVs for estimating scene geometry to be scaled using sparse points (green) part of a satellite-based Global Digital Elevation Model, yielding metric depth results}
	\label{fig_illsut}
\end{figure}

A variety of scene geometry data sources are available to aerial-based applications, each having particular strong and weak points. 
Global Digital Elevation Models (GDEM) boast high coverage but are of low horizontal resolution and are typically acquired with a low temporal frequency, increasing the possibility of 
containing outdated data. 
3D products obtained through Airborne Laser Scanning (ALS) or image-based photogrammetry processes feature significantly higher resolutions but face the same risk of stale data, in addition to reduced availability. 
Gaining traction in recent years, Neural Radiance Fields (NeRF) \cite{mildenhall2021nerf}  learn to represent a scene's geometry from a set of posed images, but are typically considered an offline solution and face challenges in large scenes. 
On-board LiDAR scanners can provide real-time, direct depth measurements at the cost of added bulk, synchronization, and calibration requirements. 
Computational solutions such as Structure from Motion (SfM) and Multi-View Stereo (MVS) use multiple images to estimate depth in real-time, but face challenges on certain textures, or lack thereof.  

Detecting and localizing dynamic objects such as pedestrians and vehicles is vital in many applications, yet all methods presented so far face challenges when modelling such objects,  ranging from significant (e.g., on-board LiDAR, NERF) to insurmountable (e.g. GDEM, ALS/photogrammetry, SfM). 
This limitation, in conjunction with others mentioned, makes deep learning based Monocular Depth Estimation (MDE) methods the best candidate solution, with the Single Image Depth Estimation (SIDE) variant being the focus of this paper. 
As estimating 3D geometry from a 2D view is an inherently ill-posed problem, MDE approaches typically produce lower quality results than other methods, but compensate by being able to provide results in real time, including for dynamic scene elements. 
Moreover, fusing additional information such as semantic or panoptic information with MDE-provided geometric data is straightforward, as both are directly derived from the RGB modality. 

Recent MDE works \cite{ranftl2020midas, yang2024depthany, yang2024depthanyv2} continue to stress the importance of high quality ground truth depth data. Unfortunately, as we highlighted in \cite{florea2022survey}, there are few real-world depth datasets covering outdoor UAV scenarios.
Self-supervised MDE methods are being developed as alternatives to supervised learning approaches to lessen the burden on data requirements. These can learn over image sequences \textit{without} the need for depth ground truth \cite{zhousfml}, but can only provide relative estimations, with an unknown scale, a limitation inherent to the training process.
The same limitation is found in recent depth foundation models Depth Anything \cite{yang2024depthany} and Depth Anything V2 \cite{yang2024depthanyv2} which follow the scale- and shift-invariant (SSI) training procedure proposed in MiDaS \cite{ranftl2020midas} in order to achieve high performance in zero-shot generalization across domains. SSI training, like in self-supervision, leads to learning a \textit{relative} depth model, instead of an \textit{absolute} one. Obtaining metric measurements remains, however, essential in most applications.

We consider robust scale recovery methods to be important milestones towards general metric depth estimation systems, as recent seminal works on relative depth estimation \cite{yang2024depthany, yang2024depthanyv2} show that decoupling scale during training can lead to impressive gains in generalization performance. Pairing such models, trained on large-scale data, with inference-time scale recovery methods designed for each application domain can enable powerful capabilities. 

To further this vision, we propose TanDepth, a novel method which capitalizes on the nearly universal coverage of Global Digital Elevation Models, such as TanDEM-X \cite{rizzoli2017tdem}, to generate metric depth maps from relative MDE models in outdoor UAV applications.
Our method involves using camera extrinsic and intrinsic parameters to project GDEM data into the camera view, with projected points taking the role of true depth anchors in the subsequent scaling step. 
Possible occlusions and alignment inaccuracies are handled by performing scaling only using ground surface points. 
We adapt the Cloth Simulation Filter \cite{zhang2016csf}, initially developed for aerial LiDAR scans, to run on 3D data from SSI relative depth estimations in order to segment ground patches, which, to the best of our knowledge, would represent its first use on MDE data. Fig. \ref{fig_illsut} graphically illustrates the TanDepth system -- the name is a lexical blend between TanDEM-X, the Global Digital Elevation Model used in our experiments and the primary output of the system, depth, symbolizing the synergistic approach we propose.

The main advantage of our proposal lies in its inherent adaptability to the 3D geometry of the observed scene, regardless of flight parameters or local geography. This enables achieving consistent high scaling performance on a variety of scenes, which, when paired with the impressive generalization capabilities of recent depth foundation models, forms a powerful and practical system for 3D reconstruction. TanDepth's adaptability stems from the integration of GDEM data as a low-resolution geometric representation of the surveyed site that can inform the scale recovery process. This comes in contrast with height-based scale recovery approaches, which cannot model terrain variations or distant observations using longer focal lengths or shallow pitch angles. Additionally, implemented as a post-processing method and designed to work even with models trained using the scale and shift invariant loss, TanDepth is agnostic to the paired depth model, and can work with the majority of relative MDEs out-of-the-box, without the need for additional training or modifications to the models.

In order to validate our method, we conduct experiments on a selection of scenes from 3 public datasets, diverse in terms of visible depth ranges, terrain profiles, vegetation, man-made structures and regional aesthetics. 
For enabling comparisons, we adapt the existing camera height-based scaling used in automotive applications to outdoor aerial environments, as well as for use with estimations coming from SSI models.
Furthermore, we release UAVid-3D-Scenes, an extension to the UAVid \cite{lyu2020uavid} dataset, to support future development of the aerial MDE field.

Our contributions presented in this paper can be summarized to:
\begin{itemize}
\item TanDepth: A novel inference-time scale recovery method based on Global Digital Elevation Model data that, when paired with modern relative depth estimators, can provide high quality zero-shot metric depth in aerial scenarios, as demonstrated through evaluations across 4 varied scenes from 3 datasets: UAVid \cite{lyu2020uavid}, UFO Depth \cite{licuaret2022ufo} and WildUAV \cite{florea2021wilduav}. The key role scaling has on generalization performance is highlighted through experiments with multiple depth models, including an absolute metric model. For evaluation, we adapt another scale recovery method based on camera height priors to function in outdoor UAV applications with scale- and shift-invariant relative depth maps.   
\item Through straightforward adaptations, we enable the use of the Cloth Simulation Filter 
 \cite{zhang2016csf} on scale- and shift-invariant monocular depth estimations for segmenting ground surfaces, outside its designed use on aerial LiDAR scans.
\item UAVid-3D-Scenes\footnote{The UAVid-3D-Scenes data, example code and further details will be made available on the \href{https://github.com/hrflr/uavid-3d-scenes}{project's GitHub page}} Dataset: We release an extension specifically focused on depth estimation and scene reconstruction tasks for the widely used UAVid \cite{lyu2020uavid} dataset. Our extension includes RGB, depth, pose and camera intrinsics information. The aim is to enable further research on aerial depth estimation tasks, by providing training and evaluation data captured in real-world scenarios. As an extension of the original UAVid dataset, it can also support future joint workloads for depth and semantic tasks.
\end{itemize}

\section{Related Work}
\textbf{Monocular Depth Estimation.} Typically framed as a supervised regression task, some of the recent advancements in MDE focus on the underlying network architecture \cite{ranftl2021dpt}, integrating camera model knowledge \cite{guizilini2023zerodepth,piccinelli2024unidepth} and optimizing learning over collections spanning multiple datasets \cite{ranftl2020midas, yang2024depthany, yang2024depthanyv2}, typically disentangling scale and providing relative estimations. 
Other lines of work focus on recasting MDE as a classification task in the form of an ordinal regression network to exploit more favorable learning traits \cite{fu2018deep, bhat2021adabins, bhat2022localbins}. This approach is proposed even solely for the task of relative to metric scaling \cite{bhat2023zoedepth}. 
Some works targeting UAVs refine the ordinal regression for longer ranges \cite{miclea2021monocular} or by using semantic guidance \cite{miclea2023dynamic} while others propose learning from virtual stereo pairs \cite{madhuanand2020deep} or reuse previous estimations \cite{fonder2021m4depth}. 
Self-supervised MDE approaches \cite{gargfirst, zhousfml, godardmono2, bian2} aim to lessen the need for annotated data during training by framing the task in the context of novel view reconstruction problems, where supervision is derived from photometric inconsistencies between real and reconstructed frames, with some works tackling the aerial domain as well \cite{madhuanand2021self, hermann2021real}.   

\textbf{Scale Recovery Methods.} Relative MDE models, irrespective of the training methodology employed, require additional subsystems in order to provide metric measurements. In the automotive domain, these can exploit information on the known, fixed camera height \cite{xue2020toward, wagstaffScale} or even auxiliary Radar data \cite{li2024radarcam} to perform scaling. Some works propose \textit{learning} scale during training, through direct supervision with pose information \cite{packnet} or using an auxiliary scale network trained on synthetic data with dense ground truth \cite{swami2022dwyc}. 
In aerial environments, \cite{pirvu2021depth} use depth computed analytically from dense optical flow and GPS-based odometry data to scale estimations from a self-supervised model in order to build a pseudo-labeled dataset that is later used to distil knowledge to a student model. Similar to works in automotive scenarios, \cite{monabench} scale depth estimations by using the UAV's height above the ground, as measured by a downward facing Time of Flight (TOF) sensor. However, the approach is limited to mainly indoor applications, due to the TOF sensor's limited range and to depth models that are scaled using a single multiplicative term.

Our proposal, unlike learning-based solutions, does not require vast amounts of data prior to deployment, as no training processes are used. The method does not depend on any other trained models (such as for optical flow or semantic segmentation), other than the relative depth model whose outputs are being scaled. This also ensures functionality in a wide variety of scenarios out-of-the-box, without having to tackle the issue of generalization. Similar to methods based on camera height \cite{monabench}, TanDepth is implemented as a post-processing step following relative depth estimation. Unlike these, however, our proposal can be applied outdoors, during flights at variable altitudes and, most importantly, is designed to integrate geometric information of the in-view terrain, ensuring operability over various terrain profiles.

\textbf{Global Digital Elevation Models.} Rapid advancements in space-based remote sensing capabilities have enabled the creation of 3D GDEMs that provide almost complete coverage of the Earth's surface. GDEM data is already used in UAV applications to enable terrain-aware mapping functions that keep flight altitude constant relative to the mapped area.
Technological advances have led to GDEMs boasting higher accuracy and resolution over time: the Shuttle Radar Topography Mission (SRTM) \cite{srtm} used Synthetic Aperture Radar (SAR), while the ASTER \cite{tachikawa2011aster} and JAXA ALOS 3D \cite{tadono2014alos} GDEMs used image-based stereoscopic reconstructions. 
For our work, we selected to use the TanDEM-X \qty[mode = text]{30}{\metre} Edited DEM which was recently released publicly. The data is part of the TanDEM-X \cite{rizzoli2017tdem} project, which uses the first space-based bistatic SAR system comprised of two near-identical satellites, providing measurements with vertical accuracy better than \qty[mode = text]{2}{\metre}, homogeneously across all land masses.

\textbf{Aerial Datasets.} Capturing high quality datasets representative for outdoor aerial perception tasks poses unique and complex challenges. Synthetic data generation solutions such as AirSim \cite{shah2018airsim} can be used to develop datasets such as Mid-Air \cite{fonder2019mid} or TartanAir \cite{tartanair2020iros}, leaving open the issue of addressing the synthetic-to-real domain gap. 
While RGB, pose and even semantic data can be obtained in a straightforward manner, capturing aerial ground truth depth maps is cumbersome and is tackled by few works. 
The WildUAV \cite{florea2021wilduav} dataset is one of the first collections providing photogrammetry-based reference depth and pose data for outdoor aerial nadir and oblique images. As depth maps obtained through photogrammetry approaches represent indirect measurements, it is more accurate to describe them as \textit{reference} depth maps instead of \textit{ground truth}.
UFO Depth \cite{licuaret2022ufo} makes available oblique UAV video collections together with accurate pose data which enable reconstructing the 3D scene. 
Recently released, the UseGeo \cite{hermann2024usegeo} dataset provides nadir images alongside LiDAR-based depth maps captured from a UAV. 
While only providing semantic annotations, the UAVid \cite{lyu2020uavid} oblique video dataset is noteworthy due to its wider adoption.

\section{Method}
\subsection{Theoretical Background and Problem Formulation}

\begin{figure*}
\footnotesize
    \centering
	\includegraphics[width=18cm]{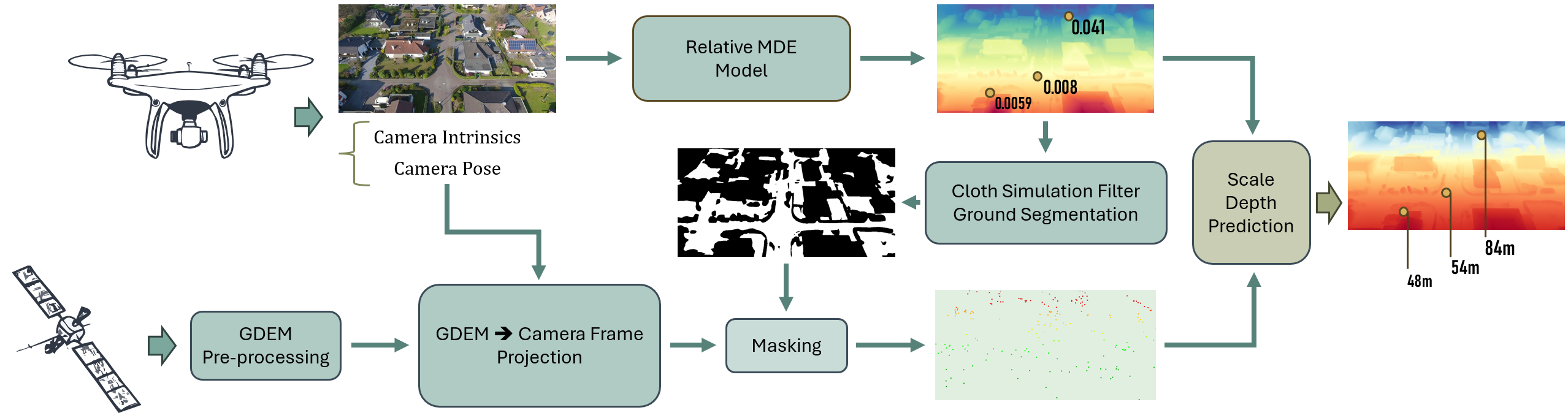}
	\caption{TanDepth processing flow. RGB, pose, intrinsics and GDEM inputs are processed to generate a sparse metric ground map which is used to scale a relative depth map estimated by a MDE model. The unscaled depth is used to generate a ground segmentation that masks out any GDEM points projected from other surfaces.}
	\label{fig:tandepth}
\end{figure*}

Traditional offline evaluation procedures for relative depth models include a scaling pre-processing step to ensure the arbitrary scale predictions match the scale (typically metric) of the reference / ground truth depth maps $D^*$ used for evaluation. 
Commonly, MDEs infer disparity $d$, which is converted to depth values as $D=1/d$. The standard depth scaling procedure computes a linear scale factor $sf$ between the median values of the two depth maps $sf=\mathrm{median}(D^*)/\mathrm{median}(D)$, which is then used to scale the predictions as: $\Bar{D}=sf*D$. Naturally, such a strategy is inapplicable to any real-world application, due to unavailability of the reference depth data. 

The trivial inference-time solution for scaling predictions is to pre-compute a series of scaling factors on frames from validation dataset (with known reference depth) and then use the average (or median) value from the series as a fixed scaling factor ($sf_{\mathrm{fixed}}$) during live operation. Clearly, this method assumes that the model's predictions exhibit scale consistency over varying datasets and conditions, which is not typically the case (especially for self-supervised models) and is an active research topic \cite{bian1}. Scale inconsistencies can be even more pronounced in aerial scenarios in which flight parameters (altitude, camera pitch, etc.) can vary and significantly modify the scene geometry in the frame. For more details on the standard scaling procedure, we direct readers to consult monodepth2 \cite{godardmono2}.

\begin{gather}
    \hat{d}=\frac{d-t(d)}{s(d)} \label{eq:ssi}\\ 
    s(d) = \frac{1}{M}\sum_{i=1}^{M}{|d-t(d)|} \qquad
    t(d)=\mathrm{median}(d)\label{eq:sandsh}
\end{gather}

The standard scaling procedure requires further adaptation for models (such as DepthAnything \cite{yang2024depthany}) trained using the scale- and shift-invariant loss, first proposed by MiDaS \cite{ranftl2020midas}. This training strategy aligns both the prediction and the reference disparity maps to have zero translation and unit scale, before computing the mean-squared error loss. For a disparity map $d$ of size $M$, the scale- and shift-invariant map $\hat{d}$ is obtained using Eq. \ref{eq:ssi}, where the scale and shift terms are computed according to Eq. \ref{eq:sandsh}. In order to realign a prediction $\hat{d}$ generated by such a model to the reference $d^*$, one can employ a closed-form solution of the least-squares criterion presented in Eq. \ref{eq:ssi_lsq}. This computes the scale and translation terms which are used to scale the prediction as $\Bar{d}=s\hat{d}+t$. This scaling procedure is compatible with other types of relative MDEs, such as those trained using self-supervision.

\newcommand{\argminD}{\arg\!\min} 
\begin{gather}
    (s,t) = \argminD_{s,t}\sum_{i=1}^{M}(s\hat{d}_i+t-d_i^*)^2 \label{eq:ssi_lsq}
\end{gather}

Use of this approach at inference-time is hampered by its dependency on the reference map $d^*$. We thus aim to find a suitable replacement that accurately captures the scale of the observed scene. First, we present an adaptation of camera height-based methods \cite{xue2020toward} that enables them to operate on scale-invariant depth maps, by replacing $d^*$ in Eq. \ref{eq:ssi_lsq} with $h^*$, computed based on known height and camera pitch. This is followed by our proposed method, TanDepth, which computes a sparse metric ground map $\tilde{g}^*$ from GDEM data which more accurately captures the local scene's geometry.

\subsection{Adapting Camera Height Scaling for Scale-invariant MDEs} \label{sect:camheight}

The authors of \cite{xue2020toward} present a scale recovery method designed for automotive applications, where cameras are typically placed at fixed, known heights on the vehicle's body. 
This prior is exploited to compute a scaling factor which places the road surface detected in the depth map at the correct height, thus scaling the results of the self-supervised model. 
We adapt this method for both the aerial domain and for scale and shift invariant models in order to use it as a benchmark of scaling performance.

Estimated depth information is first back-projected to 3D space using camera intrinsics, after which, in aerial scenarios, the ground surface must be aligned to the XY-plane, by rotating the cloud to compensate for the camera pitch angle. 
The fixed height prior must also be adapted for use in UAVs, by replacing it with a dynamic Above Ground Level (AGL) value $h_{\mathrm{agl}}$. This can be provided directly by devices such as Radar altimeters or can be computed based on the true Mean Sea Level (MSL) altitude and a DEM of the flight area.
Surface normals are computed for all depth map pixels in order to select ground points $gp_{i}$ (with normals close to ideal ground normal). In our implementation we augment the selection by subsequent filtering using CSF segmentation (see Section \ref{sect:csf}). For each selected ground point, its height $h_{gp}^{i}$ can be retrieved as the vertical axis component in the back-projected, pitch-aligned 3D point cloud.

Originally, the scaling factor $sf$ is computed by correlating the median ground point height with the known camera height as: $sf=h_{\mathrm{agl}}/\mathrm{median}(h_{gp})$. This works for self-supervised models, but fails for scale invariant ones, as replacing the inferred, invariant disparity $\hat{d}$ with its corresponding height map $h_{gp}$ (and $d^*$ with $h_{\mathrm{agl}}$) in Eq. \ref{eq:ssi_lsq} leads to the naive solution $(s=0, t=h_{\mathrm{agl}})$, which cancels out the inferred disparity. 
To solve this and enable the method's use with a larger variety of depth models, we propose performing the scaling operation directly in the \textit{disparity} space of the network's output instead of the \textit{height} space. 

\begin{figure*}
    \scriptsize
    \centering
	\includegraphics[width=18cm]{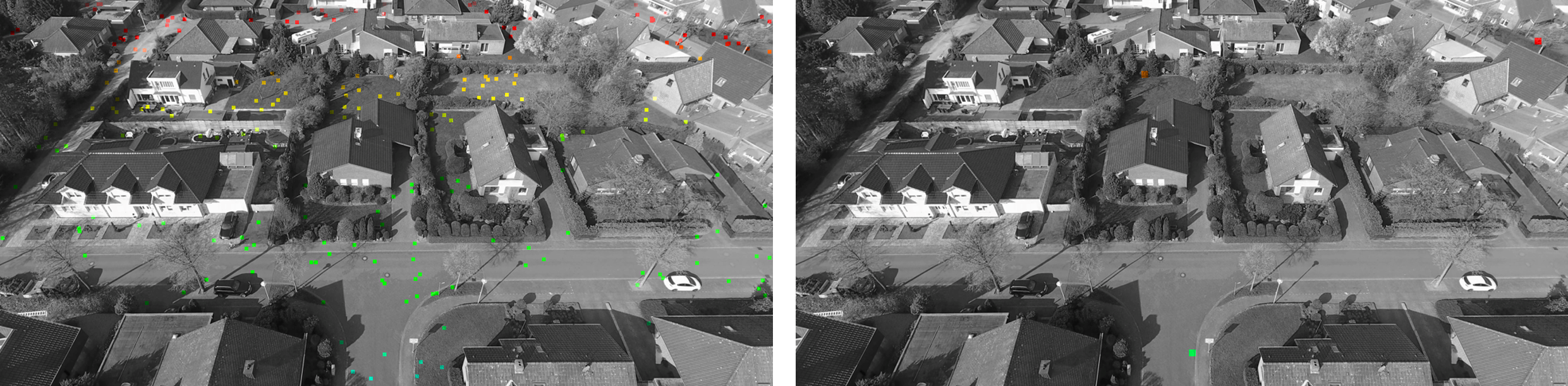}
	\caption{TanDEM-X GDEM Projection. Example of GDEM points projected to the image. Left frame shows points from the densified GDEM, right shows raw points from TanDEM-X; in both cases, non-ground points are masked out}
	\label{fig:tdem_projs}
\end{figure*}

Our adaptation builds a height-based disparity map $h^{*}$ that models the ground as a flat planar surface. For each row $r$ in $h^{*}$ we compute the distance a ray passing through $r$ travels until intersecting the ground. This can be computed as $dist_{r} = h_{\mathrm{agl}}/cos(\alpha_{r})$, where $\alpha_{r}$ is the angle of the ray relative to the ground, taking into account camera pitch and Field of View (FoV) -- an angle of $0^{\circ}$ corresponds to rays parallel to the ground plane, while those orthogonal feature an angle of $90^{\circ}$. The $h^{*}$ map, containing disparities corresponding to each row's  $dist_{r}$, replaces the reference disparity map $d^*$ in Eq. \ref{eq:ssi_lsq}, yielding $(s,t)$.

\subsection{TanDepth}

Our proposed scale recovery method aims to solve the principal limitation of the camera height-based method in aerial scenarios, namely its simplified modelling of the ground as a single planar surface. While this operational assumption is valid in automotive applications where the camera moves closely above the road surface, this is not the case for UAV operations. 
Another key difference between automotive and aerial scenarios which affects camera height based methods is the distance to the closest ground points that are visible in the frame. The distance, typically very short in driving applications, is strongly correlated with height, focal length and pitch angle. Larger distances can lead to inaccurate use of the AGL data, which is measured (or computed) directly under the UAV.

TanDepth addresses all such limitations by integrating Global Digital Elevation Model data as a low resolution geometric model of the observed scene. Using camera pose information and intrinsic calibration, GDEM data is projected to the image frame, serving as a terrain-aware prior generated  specifically for the current viewpoint. 
More specifically, we compute a sparse ground map $\tilde{g}^*$ where each pixel $i$ to which a GDEM point is projected stores the disparity $g_{i}$ corresponding to the Z-axis distance to the point. 
This map can replace the reference map $d^*$ in Eq. \ref{eq:ssi_lsq} and be used to determine the factors $(s,t)$. The method is comprised of a pre-processing step which prepares the DEM data, followed by the occlusion-aware projection to the frame and the masking of non-ground pixels using a ground segmentation based directly off the inferred 3D data, all detailed in the following sections and illustrated in the block diagram in Fig. \ref{fig:tandepth}.

\subsubsection{Pre-flight Setup}

As our system operates within a metric coordinate system, we first reproject the TanDEM-X GDEM data from its original WGS84 coordinate reference system (CRS), based on geodetic coordinates (latitude and longitude), to the Universal Transverse Mercator system that uses metric coordinates within pre-defined UTM zones. The flight location determines which zone will be used (e.g. UTM Zone 32N for UAVid Germany data). Large numerical values on the horizontal (XY) axes can lead to lower accuracy in the  projection process, an effect which is alleviated by translating the original coordinates to a local coordinate system centered in the area of interest, using a global shift vector (similar to how modern point cloud processing software operate \cite{CloudCompare_software}).
In our tests, all pre-processing steps were performed using the QGIS \cite{QGIS_software} and CloudCompare \cite{CloudCompare_software} open-source software suites.

In cases where absolute height (WGS84 altitude) data is not available, an additional Altitude Sync step must be performed in order to correlate relative height information with the vertical datum of the GDEM. This is usually required only in cases of offline processing, for public datasets which lack absolute height information -- most professional and consumer-grade UAVs now directly report absolute height.   
The Altitude Sync is achieved using a frame in which the drone's relative height is known (e.g. while stationary on the ground, hovering at a fixed height, etc.) and recording the altitude of the GDEM point horizontally closest to the position used as the drone's reference height. This vertical shift value can then be added to the relative altitude of the UAV during flight.

Our initial tests revealed the sparseness of TanDEM-X can negatively impact the stability and scaling performance of the system. In the most challenging scenarios, defined by low flight altitude combined with camera pitch angles close to nadir, few GDEM points are visible in the image. In some cases, no ground GDEM points can be projected, making scale recovery impossible.  
To mitigate this, we densify the GDEM data by first computing a Delaunay 2.5D triangulation based on the original (raw) points followed by randomly sampling a new, denser point cloud off it. This not only ensures the successful processing of all frames, but also leads to improved scaling performance. 
We use a sampling density empirically set to \qty[per-mode = symbol]{0.05}{\points\per\square\metre} -- an ablation study for this value is presented in Section \ref{sect:ablationstudies}. The densification procedure increases the number of points used in the scale recovery process from an average of 9 pts/image (using the raw GDEM data) to 335 pts/image (as measured on the UFO Depth Oveselu scene).
A comparison of projections based on the densified and original GDEM data is shown in Fig. \ref{fig:tdem_projs}.

An alternative method would involve performing the densification using only ground GDEM measurements, instead of the entire raw GDEM data, to ensure even higher accuracy. As the classification of the points is not known ahead of time, this strategy would have to be performed for each frame, after points are projected and non-ground points are rejected based on the ground segmentation. This, however, would bring about several limitations: added  computational overhead at runtime for interpolating in 3D space and a second round of projections, the risk of few or no initial GDEM points to interpolate from and the possibility of ignoring true ground points which are occluded by other structures.    

\begin{figure*}
    \scriptsize
    \centering
	\includegraphics[width=18cm]{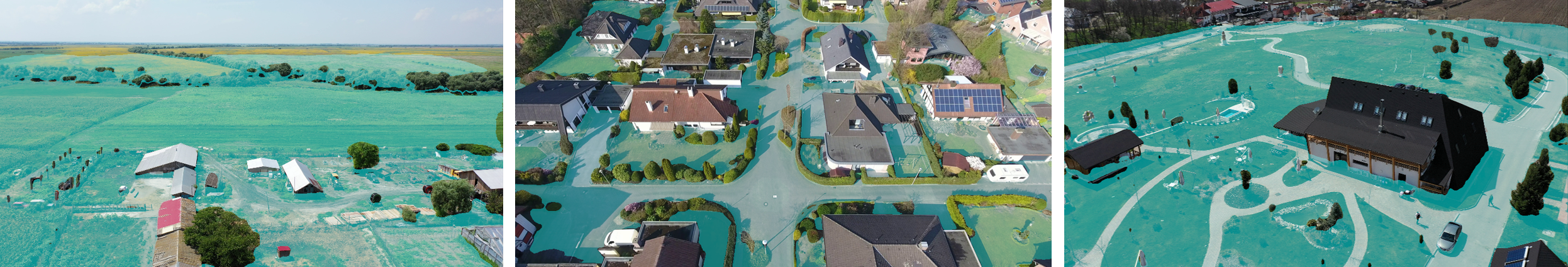}
	\caption{CSF Ground Segmentation. Results of the ground segmentation based on adapted CSF, overlayed as a turquoise layer over input images from Chilia, Germany and Oveselu scenes, respectively.}
	\label{fig:csf_segs}
\end{figure*}

\subsubsection{Occlusion-aware GDEM Data Projection} \label{sect:projection}

First, the camera's extrinsic parameters $R_{c}, T_{c}$ are used to transform the GDEM points from world coordinates to the local camera coordinate system, after which the intrinsic matrix $K$ is used to project them onto the image plane. 
Points with positive Z distance in the camera coord. sys. and which are mapped inside the image bounds are kept, and their Z coordinate is stored in the sparse map $\tilde{g}^*$ which will be used in the scaling process. 
Conflicts in which multiple points are mapped to the same pixel are resolved by keeping only the closest, occluding measurement.

This approach cannot solve all occlusion cases due to the sparseness of the GDEM data, even after densification -- points belonging to surfaces occluded in the image can still be projected if not directly behind another point from the occluding surface. 
Comprehensive solutions for this hidden point removal problem are available \cite{katz2007direct}, yet we opted for a simpler solution in our implementation as only a rough filtering is sufficient for the scaling task: a projected point $g_{i}$ is marked as occluded and discarded if its neighbourhood contains any points that are closer than $g_{i}$ to the camera, as they signal a possible occluding object. The neighbourhood is defined as a window $W$ of a wider aspect ratio ($h=w \div 2, w=7$ in our tests) as the aerial viewpoint typically makes depth values more consistent across (non-occluded) image rows. A threshold $th_{occ}$ is used to reduce the number of false rejections, by only considering $g_{i}$ occluded if it is farther than the closest point in the window by $th_{occ}$ units. We adapt this mechanism for aerial use cases by defining the threshold in relation to the evaluated depth $th_{occ}=0.04*g_{i}$, as the magnitude of depth discontinuities increases with distance.

Before using the sparse metric ground map $\tilde{g}^*$ in Eq. \ref{eq:ssi_lsq}, it must be constrained to feature only true ground points within the depth range of interest. 
To this end, we use two masks: a 2D ground segmentation mask $M_{CSF}$ computed using the Cloth Simulation Filter and a range mask $M_{Rng}$ which masks out points outside the target depth range, similar to how reference depth maps are masked in standard evaluation practice.
Finally, as the scale and shift computation takes place in disparity space, the depth values in $\tilde{g}^*$ are inverted. 

\subsubsection{Adapted Cloth Simulation Filter for Scale-invariant Ground Segmentation} \label{sect:csf}

Natural and man-made surface features can negatively impact the vertical accuracy of TanDEM-X data \cite{wessel2018tdemaccuracy}. Such features can also affect the projection and scaling process, through occlusions and presence of outdated structures in the scene. To address this, our scaling solution is designed to reject any non-ground or occluded GDEM points by segmenting visible terrain patches in the input image. One possible solution for achieving this is using a deep learning based semantic segmentation model trained on annotated aerial data. This, however, would add significant costs in terms of data, training and deployment complexity of the complete system. We avoid these drawbacks by adopting the Cloth Simulation Filter \cite{zhang2016csf} (CSF) method for the segmentation task, which was first developed for extracting Digital Terrain Models (DTMs) from airborne LiDAR data. CSF works by covering an inverted copy of the 3D scene with a simulated cloth and using its interactions with the 3D points to model the terrain surface and segment the point cloud (PC). 

One of CSF's strengths is the relatively low number of parameters requiring tuning, compared to other computational methods. In our system's case however, selecting and adjusting the cloth resolution and class threshold parameters to each frame represents a challenge, as their values are expressed in the same units as the input point cloud. As the initial PC on which the CSF is applied has an unknown scale, a circular dependency is created, made worse by the significant warping of PCs obtained from scale-invariant MDE networks. 

To solve this, we propose a straightforward adaptation that enables the use of the Cloth Simulation Filter for segmenting ground patches from MDE-generated inputs. As CSF operates on 3D point clouds, first, the disparity inferred by the network must be converted to depth measurements. These are then back-projected using the inverse intrinsic matrix, \(K^{-1}\) to form the initial PC. In order to reduce warping inherent to scale-invariant estimations, we perform an initial scaling as per Eq. \ref{eq:ssi_lsq} using a pair of constant parameters, $\overline{s}$ and $\overline{t}$. This represents, in effect, a type of fixed scaling, in which the values of the two parameters are precomputed on a validation dataset for which reference depth is available (UAVid Germany Val, in our case). Thus, we are able to compute a rough estimation for a metric depth map, denoted $\overline{D}$. This initial scaling is only necessary when using MDEs trained using the scale and shift invariant loss.

\begin{gather}
    D^*_{central} = \frac{h_{\mathrm{agl}}}{cos(\mathrm{pitch_{cam}})} \label{eq:tgtdist}\\
    \overline{D}_{central} = \mathrm{centralRowsMedian}(\overline{D}) \\
    cf = \frac{D^*_{central}}{\overline{D}_{central}} \label{eq:paramfactor}
\end{gather}

While simple, scaling using the fixed values can produce poor results, especially if the target scenario differs significantly from the one in the validation set used to set the parameters' values. As such, we introduce a secondary refinement step which computes an adjustment factor value $cf$ based on the camera pitch and height above ground of the UAV (AGL, which can be derived from GDEM data and absolute height information, if not directly provided by the platform). 
Specifically, we use the height and pitch values to estimate the distance $D^*_{central}$ to the expected ground surface in the central rows of the image, following Eq. \ref{eq:tgtdist}, similar to how we build $h^*$ in Section \ref{sect:camheight}). This distance estimate is then used to compute $cf$ by dividing it to the median depth value in the central rows of the rough depth map $\overline{D}$, following Eq. \ref{eq:paramfactor}. In our experiments we use the central 35 rows of the image for computing $cf$.

Finally, the CSF algorithm can be applied on the point cloud back-projected from $\overline{D}$, using the parameters $1.5/cf$ and $0.5/cf$ for the cloth resolution and class threshold, respectively. For UAVid Germany, we use alternate values of $0.5/cf$ and $1.25/cf$, better suited for the more cluttered scene. As each depth map pixel has a correspondent in the 3D point cloud, it is trivial to convert the CSF output (segmented PC) into a 2D segmentation mask, $M_{CSF}$. This is then used to reject any GDEM points projected onto or behind of non-terrain image patches. Examples of generated masks are depicted in Fig. \ref{fig:csf_segs}.

Algorithm \ref{alg:main} presents the main components of the proposed method in pseudo-code format: the main TanDepth function represents the entry point of the algorithm featuring calls to the CSF segmentation and GDEM data projection functions, followed by computation of the scaling parameters (using least-squares criterion as defined by MiDaS \cite{ranftl2020midas}) and their use on the relative depth image. The CSF function acts as a wrapper for the official CSF module, performing our proposed adaptation, while the projection function transforms the 3D data based on the camera's extrinsic ($Pose$) and intrinsic parameters ($K$), while also performing the occlusion handling pass. 

\begin{algorithm}
\caption{TanDepth Algorithm in Pseudo-code}\label{alg:main}
\begin{algorithmic}
\State $RelDisp \gets \Call{RelativeDepthModel}{RGB}$
\State $AbsDepth \gets \Call{TanDepht}{RelDisp, K, Pose, GDEM}$
\State 

\Function{TanDepth}{RelDisp, K, Pose, GDEM}
    \State $gnd \gets \Call{CSF}{RelDisp, Pose.Pitch, Pose.AGL, K}$
    \State $refPts \gets \Call{Project}{K, Pose, GDEM}$
    \State $gndRefPts \gets refPts \cap gnd$
    \State $gndDisp \gets 1/gndRefPts$
    \State $s, t \gets \Call{LeastSquaresMin}{RelDisp, gndDisp}$
    \State $scaledDisp \gets s \cdot RelDisp + t$
    \State $metricDepth \gets 1/scaledDisp$
    \State \Return metricDepth
\EndFunction
\State

\Function{CSF}{RelDisp, Pitch, Height, K}
    \LComment{$\Bar{s}, \Bar{t}$ determined on validation set}
    \State $roughDpt \gets 1 / (\Bar{s} \cdot RelDisp + \Bar{t})$
    \State $pcl \gets \Call{BackProject}{roughDpt, K^{-1}}$
    \State $estCenter \gets \Call{CentralRowsMedian}{roughDpt}$
    \State $expectedCenter \gets Height/\cos(Pitch)$
    \State $cf = expectedCenter/estCenter$
    \State $CSFmodule.cloth\_res\gets1.5/cf$
    \State $CSFmodule.class\_th\gets0.5/cf$
    \State $groundSeg \gets \Call{CSFmodule}{pcl}$
    \State \Return groundSeg
\EndFunction
\State 

\Function{Project}{K, Pose, GDEM}
    \State $GDEMproj \gets K \times Pose \times GDEM$
    \State $pts \gets GDEMproj / GDEMproj.Z$
    \State $pts.Z \gets GDEMproj.Z$
    \State $mask \gets (pts.Z \geq 0) \cap (pts.X<W) \cap (pts.Y<H)$
    \State $pts \gets pts \cap mask$
    \State $ptsImg \gets \Call{zeros}{W, H}$
    \For{$p \in pts$}
        $ptsImg[p.X,p.Y]\gets p.Z$
    \EndFor
    \State $sz\gets(5,7)$
    \State $th \gets 0.04 \cdot ptsImg$
    \For{$p \in pts$}
        \State $min \gets \Call{WindowMin}{ptsImg,p,sz}$
        \If{$p.Z-min>th[p.X, p.Y]$}
            \State $ptsImg[p.X,p.Y]\gets0$
        \EndIf
    \EndFor
    \State \Return ptsImg
\EndFunction
\State 

\end{algorithmic}
\end{algorithm}

\subsection{The UAVid-3D-Scenes Dataset} \label{sect:UAVid-3D-Scenes}

\begin{figure*}[htbp]
  \centering
  \begin{subfigure}{\linewidth}
    \scriptsize
    \centering
    \includegraphics[width=18cm]{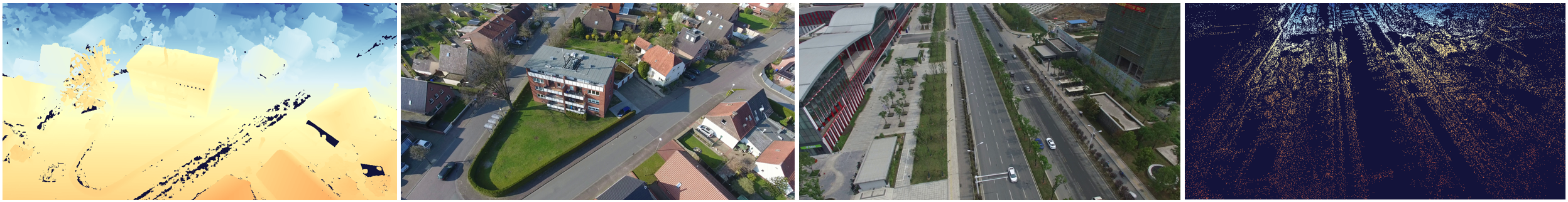}
    \caption{}
	\label{fig:UAVid-3D-Scenes}
  \end{subfigure}
  
  \vspace{0.5em}
  
  \begin{subfigure}{\linewidth}
    \scriptsize
    \centering
    \includegraphics[width=18cm]{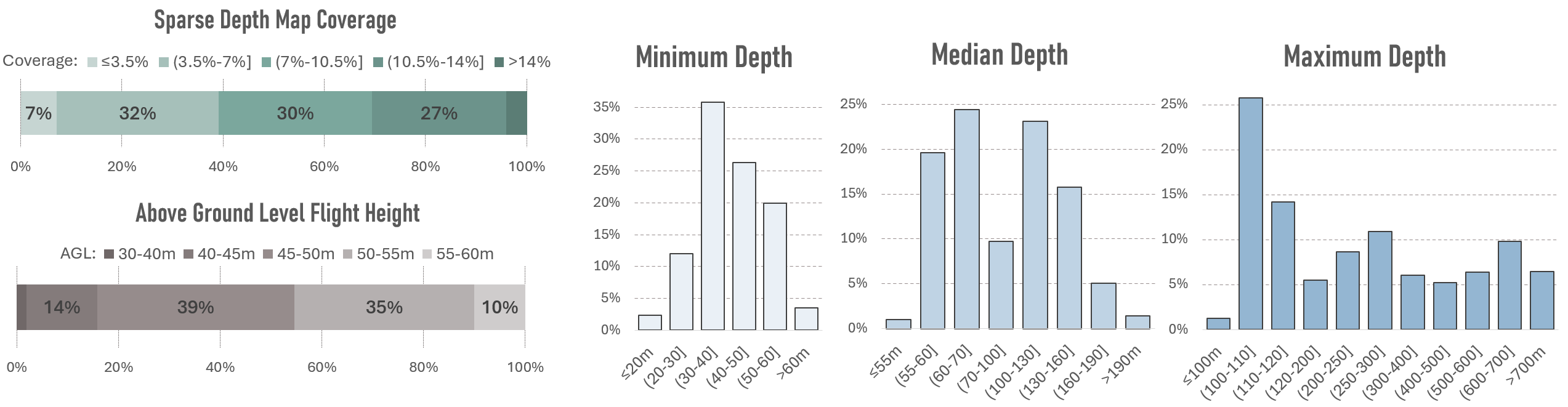}
    \caption{}
	\label{fig:dataset-analysis}
  \end{subfigure}
  
  \caption{UAVid-3D-Scenes. (a) Examples of the dense (left) and sparse (right) depth reconstructions for UAVid scenes recorded in Germany and in China, respectively. (b) Distribution of key metrics in UAVid-3D-Scenes: Coverage (\% of reconstructed pixels in a frame) of the sparse depth maps, AGL flight height and the per frame minimum, median and maximum depths across the entire dataset.}
  \label{fig:combined}
\end{figure*}

To enable further development of the aerial MDE field, we release a depth-centric extension to the UAVid \cite{lyu2020uavid} dataset we name UAVid-3D-Scenes. Its goal is to provide necessary depth and pose reference data for developing advanced solutions addressing various geometric tasks in aerial contexts. By building on the existing UAVid dataset, which was designed solely for semantic segmentation tasks, UAVid-3D-Scenes can also enable future work on joint tasks involving scene geometry and semantic information, beyond the immediate use for MDE applications. UAVid-3D-Scenes also stands out by virtue of the scene types covered, providing depth information for real-world data in urban and suburban environments, as originally captured in UAVid in the form of oblique videos -- coming in contrast with synthetic datasets \cite{fonder2019mid}, those captured over rural areas \cite{florea2021wilduav} or those captured as nadir mapping image sets \cite{hermann2024usegeo}.

We use the robust COLMAP \cite{colmap1, colmap2} software to generate the 3D reconstructions, which models the process as a pipeline with four main components: feature extraction, feature matching, sparse reconstruction (Structure from Motion) and dense reconstruction (Multi View Stereo). We release a total of 5517 samples for which the undistorted RGB image and the depth map image is provided. 
Out of these, 2985 frames (sampled from videos at rates of 2 and 5 FPS in Germany, 2 FPS in China) have an associated dense depth label $C$ generated by running the complete pipeline, including the Multi View Stereo (MVS) step, while the rest (sampled at 2 FPS) are released with associated sparse depth labels $C^*$. These are obtained by using the Occlusion-aware projection mechanism presented in Section \ref{sect:projection} to project 3D points resulting from the Structure from Motion (SfM) reconstruction stage in COLMAP. By sampling across the entirety of UAVid \cite{lyu2020uavid}, we ensure our extension matches the diversity of the original scenarios. The assets are provided at a resolution of $512\times1024$px.

Originally, the UAVid \cite{lyu2020uavid} dataset was released as a collection of 42 fixed length (901 frames) video sequences, 9 of which were recorded in Germany and the rest in China. We rearrange the sequences in order to perform the 3D reconstructions at \textit{scene} level, where a scene can be composed of one or more sequences recorded over the same area. This adds more context to the data and can help the reconstruction through loop closures. We thus group the original sequences into 12 scenes, based on manual matching of the landmarks in the original videos. 

As other works have noted \cite{madhuanand2021self}, the Germany and China subsets exhibit core differences in terms of scene content and even recording parameters, impacting the performance of 3D reconstruction tasks. As such, dense reconstructions were successful for both scenes in Germany, while the large number of dynamic elements in China allowed only one scene to be densely reconstructed. The challenging conditions affected the sparse reconstructions as well, prompting us to manually clean-up the point clouds, in addition to using CloudCompare's \cite{CloudCompare_software} Statistical Outlier Removal filter for the purpose of ensuring adequate data quality. Additional manual masking of image horizon regions was required for dense maps in scenes from China, in order to remove false reconstructions. Examples of the dense and sparse reconstructions are depicted in Fig. \ref{fig:UAVid-3D-Scenes} -- dark blue pixels represent patches lacking depth information.

Multiple factors inherent to the locations and selected flight profiles play important roles in the training process of depth estimation systems. Following an analysis of the developed dataset, Fig. \ref{fig:dataset-analysis} presents the distribution of key metrics that impact geometric tasks. First, we can observe that 57\% out of the total sparse depth maps feature a coverage (number of reconstructed pixels in relation to depth map size)  between 7\% and 14\% -- for reference, a 10\% coverage equates to $\approx52000$ points. In contrast, 89.7\% of dense depth maps provide over 90\% coverage. The flights feature only small variations in terms of Above Ground Level height, with the majority of frames recorded in the \qtyrange{40}{50}{\metre} interval. 

AGL's stability carries over to the minimum values encountered in each depth map, which rarely lie outside the \qtyrange{20}{60}{\metre} range, with over a third of values between \qtyrange{30}{40}{\metre}. The maximum values, on the other hand, feature significantly more variation, especially for scenes recorded in China. Here, after excluding the top 1\% values in each frame which can originate from very small patches on the horizon, a peak in the distribution can be observed for the \qtyrange{100}{110}{\metre} range, owing primarily to data from Germany. Two thirds of frames exhibit maximum values below \qty{300}{\metre}. As the larger max values are primarily the result of varying pitch angles (with little variation in focal length and AGL), the median depth values exhibit higher stability and lower overall values, with 44\% of values contained in the \qtyrange{55}{70}{\metre} interval, and another 39\% between \qtyrange{100}{160}{\metre}. Users of the dataset should take into account the selection based on regionality made during development of the dense and sparse sets, with all German frames integrated in the dense set.

We perform an evaluation of the generated 3D reconstruction in Germany by comparing it with publicly available aerial LiDAR scans of the surveyed area. Due to the large time difference between the UAVid and LiDAR acquisitions (approximately 4 years), the evaluation is carried out for rooftop surfaces which present the greatest stability (vegetation, dynamic elements, etc. are significantly changed between the two acquisitions). For the $\approx95000$ rooftop points in the LiDAR scan the difference to the COLMAP reconstruction exhibits a mean value of \qty{0.211}{\metre} and a median of \qty{0.149}{\metre}, with 80\% of points having an error below \qty{0.35}{\metre}. These results show the error between the COLMAP reconstruction and LiDAR point cloud to be a full order of magnitude below that observed during MDE evaluations, confirming the former’s applicability during evaluations.

We additionally provide camera extrinsic (pose) data for each sample along with camera intrinsics, both obtained through the sparse SfM reconstruction process. Metric scaling of the reconstructed models is derived differently for the two locations: in Germany scale and global position was recovered by aligning the reconstructions (through scaling, rotation and translation operations) with public aerial LiDAR 3D scans, using initial alignment by manually defining point correspondences, followed by refinement using the Iterative Closest Point algorithm. As no such LiDAR scans were publicly accessible for scenes recorded in China, the scale was recovered by manually correlating sizes of multiple reconstructed landmarks in each scene (e.g. width of roads, buildings, etc.) and their metric size public 2D map data.

As part of their work on self-supervised MDE for aerial environments the authors of \cite{madhuanand2021self} have released a small subset (3 sequences) of depth reconstructions for the UAVid \cite{lyu2020uavid} dataset, under the name of UAVid-Depth. Our proposal differs both in scope (covering the entire UAVid dataset \cite{lyu2020uavid}) as well as in terms of contents (providing camera intrinsics, depth maps as well as pose information at scene level, with global position estimations for the German sequences).

\section{Experiments}

\begin{table*}
\centering
\scriptsize
\caption{Depth Anything \cite{yang2024depthany} with different scaling approaches. TanDepth outperforms Fixed and Camera Height online scaling methods across the tested datasets (best results in bold), approaching the performance of offline reference scaling (in italic). ZoeDepth is a metric depth estimation model trained on UAVid}
\begin{tabular}{crcccccccccc}
\toprule
\multirow{2}{*}{Scene}                                                                  & \multirow{2}{*}{Scaling} & \multicolumn{4}{c}{Lower is better $\downarrow$}                                                                      & \multicolumn{6}{c}{Higher is better $\uparrow$} 
\\ \cmidrule(lr){3-6} \cmidrule(lr){7-12}
                                                                                        ~ &  ~                        & \multicolumn{1}{c}{AbsRel} & \multicolumn{1}{c}{SqRel} & \multicolumn{1}{c}{RMSE} & \multicolumn{1}{c}{LogRMSE} & \multicolumn{1}{c}{${\delta}_1$} & \multicolumn{1}{c}{${\delta}_2$} & \multicolumn{1}{c}{${\delta}_3$} & \multicolumn{1}{c}{$\Bar{\delta}_1$} & \multicolumn{1}{c}{$\Bar{\delta}_2$} & \multicolumn{1}{c}{$\Bar{\delta}_3$} \\ \hline
\multirow{5}{*}{\begin{tabular}[c]{@{}c@{}}UAVid\\ Germany\end{tabular}} & Fixed                    & 0.062                        & 0.424                       & 4.693                    & 0.072                         & 0.983                  & 0.999                  & 1.000                  & 0.265                    & 0.494                    & 0.714                    \\
                                                                                      & Cam. Height              & 0.039                        & 0.262                       & 3.854                    & 0.055                         & 0.988                  & 0.999                  & 1.000                  & 0.519                    & 0.763                    & 0.878                    \\
                                                                                      & ZoeDepth*                & 0.041                        & 0.218                       & 3.659                    & 0.054                         & 0.992                  & 0.999                  & 1.000                  & 0.358                    & 0.720                    & 0.899                    \\
                                                                                      & TanDepth                 & \textbf{0.031}                       & \textbf{0.143}                       & \textbf{2.800}                    & \textbf{0.044}                         & \textbf{0.993}                  & \textbf{1.000}                  & \textbf{1.000}                  & \textbf{0.523}                   & \textbf{0.863}                    & \textbf{0.944}                    \\
                                                                                      & \textit{Reference}              & \textit{0.027}               & \textit{0.133}              & \textit{2.720}           & \textit{0.041}                & \textit{0.994}         & \textit{1.000}         & \textit{1.000}         & \textit{0.654}           & \textit{0.879}           & \textit{0.938}           \\ \noalign{\vskip -5pt} \\
\multirow{5}{*}{\begin{tabular}[c]{@{}c@{}}UFO Depth\\ Oveselu\end{tabular}}               & Fixed                    & 0.218                        & 6.966                       & 25.147                   & 0.290                         & 0.485                  & 0.800                  & 0.972                  & 0.076                    & 0.148                    & 0.213                    \\
                                                                                      & Cam. Height              & 0.112                        & 2.097                       & 12.233                   & 0.124                         & 0.885                  & 0.996                  & 1.000                  & 0.184                    & 0.336                    & 0.462                    \\
                                                                                      & ZoeDepth*                & 0.171                        & 2.933                       & 14.866                   & 0.185                         & 0.707                  & 0.998                  & 1.000                  & 0.069                    & 0.140                    & 0.212                    \\
                                                                                      & TanDepth                 & \textbf{0.048}                       & \textbf{0.344}                      & \textbf{5.498}                   & \textbf{0.061}                        & \textbf{0.997}                 & \textbf{1.000}                 & \textbf{1.000}                 & \textbf{0.330}                   & \textbf{0.592}                   & \textbf{0.784}                    \\
                                                                                      & \textit{Reference}              & \textit{0.040}               & \textit{0.285}              & \textit{5.057}           & \textit{0.053}                & \textit{0.997}         & \textit{1.000}         & \textit{1.000}         & \textit{0.430}           & \textit{0.701}           & \textit{0.849}           \\ \noalign{\vskip -5pt} \\
\multirow{5}{*}{\begin{tabular}[c]{@{}c@{}}UFO Depth\\ Chilia\end{tabular}}                & Fixed                    & 0.425                        & 28.214                      & 67.528                   & 0.614                         & 0.036                  & 0.295                  & 0.680                  & 0.001                    & 0.002                    & 0.004                    \\
                                                                                      & Cam. Height              & 0.052                        & 0.675                       & 9.078                    & 0.066                         & 0.991                  & 0.999                  & 1.000                  & 0.347                    & 0.602                    & 0.774                    \\
                                                                                      & ZoeDepth*                & 0.116                        & 3.072                       & 23.232                   & 0.151                         & 0.850                  & 0.998                  & 1.000                  & 0.095                    & 0.201                    & 0.319                    \\
                                                                                      & TanDepth                 & \textbf{0.040}                        & \textbf{0.442}                     & \textbf{7.520}                   & \textbf{0.055}                      & \textbf{0.994}                & \textbf{0.999}                 & \textbf{1.000}                  & \textbf{0.427}                    & \textbf{0.715}                    & \textbf{0.872}                    \\
                                                                                      & \textit{Reference}              & \textit{0.035}               & \textit{0.362}              & \textit{7.064}           & \textit{0.049}                & \textit{0.995}         & \textit{0.999}         & \textit{1.000}         & \textit{0.496}           & \textit{0.779}           & \textit{0.904}           \\ 
                                                                                      \noalign{\vskip -5pt} \\
\multirow{5}{*}{WildUAV}                                                              & Fixed                    & 0.345                        & 7.052                       & 18.334                   & 0.304                         & 0.295                  & 0.913                  & 0.997                  & 0.007                    & 0.015                    & 0.026                    \\
                                                                                      & Cam. Height              & 0.207                        & 9.483                       & 10.981                   & 0.178                         & 0.828                  & 0.959                  & 0.966                  & 0.116                    & 0.230                    & 0.345                    \\
                                                                                      & ZoeDepth*                & 0.502                        & 14.876                      & 27.551                   & 0.415                         & 0.062                  & 0.673                  & 0.977                  & 0.001                    & 0.001                    & 0.003                    \\
                                                                                      & TanDepth                 & \textbf{0.072}                       & \textbf{1.269}                       & \textbf{4.106}                   & \textbf{0.085}                       & \textbf{0.962}                & \textbf{0.996}                 & \textbf{0.997}                 & \textbf{0.322}                  & \textbf{0.531}                   & \textbf{0.670}                    \\
                                                                                      & \textit{Reference}              & \textit{0.041}               & \textit{0.157}              & \textit{2.560}           & \textit{0.052}                & \textit{0.995}         & \textit{1.000}         & \textit{1.000}         & \textit{0.448}           & \textit{0.708}           & \textit{0.846}           \\ 
                                                                                      \bottomrule
\end{tabular}
\footnotesize
\label{tab:mainresults}
\end{table*}

\subsection{Datasets, Evaluation \& Implementation Details}

We use Depth Anything \cite{yang2024depthany} as the baseline relative depth model using its default inference setup, as we consider it an ideal representative for relative MDE models with strong generalization capabilities. All tests are performed at the $512\times1024$ resolution (both input image and reference depth map) on undistorted images. The evaluation reference depth maps were generated using the dense reconstruction pipeline in COLMAP \cite{colmap1, colmap2}. The TanDEM-X \cite{rizzoli2017tdem} Global Digital Elevation Model used in the scale recovery process was interpolated to a density of \qty[per-mode = symbol]{0.05}{\points\per\square\metre}. The code was developed in Python and uses the official implementation of the Cloth Simulation Filter for the language.

For evaluating TanDepth we employ 542 frames from the proposed UAVid-3D-Scenes dataset, part of the captures in Germany which feature a flat terrain profile. 
From the UFO Depth \cite{licuaret2022ufo} dataset, we use the Chilia (flat terrain, 301 frames) and Oveselu (hilly terrain, 299 frames) scenes in the evaluation for which we perform dense 3D reconstructions using COLMAP \cite{colmap1, colmap2} that are then scaled by aligning the reconstructed poses with the corresponding GPS data. We measure the error of the two scene reconstructions by comparing the dimensions of various landmarks in the scaled 3D point clouds with public map data. The average observed error was \qty{0.182}{\metre} for Chilia, and \qty{0.173}{\metre} for Oveselu.
From WildUAV \cite{florea2021wilduav}, we use 265 oblique images from the hilly "scene1" capture, which features RTK positioning data. 
Evaluation is carried out for the \qtyrange{30}{150}{\metre} range which covers most of the frame for these captures, with an extended range of \qtyrange{50}{250}{\metre} range used for Chilia, where the shallower pitch angle leads to longer in-frame distances. 
Global positioning information for each frame was directly available for UFO Depth \cite{licuaret2022ufo} and WildUAV \cite{florea2021wilduav}, while for UAVid \cite{lyu2020uavid} Germany it was recovered through registration with public aerial LiDAR scans of the area (see Section \ref{sect:UAVid-3D-Scenes}).

Training an absolute depth model can be an alternative to a system combining a relative model with an inference-time scale recovery method. This involves using a supervised learning approach on sufficient (-ly diverse) metric reference depth data, a difficult task for aerial tasks.
To further evaluate the effectiveness of our proposed scaling method, we want to compare its performance with a metric depth model adapted for the aerial domain, following the same recipe proposed in \cite{yang2024depthany} for the automotive and indoor domains. 

Specifically, we use ZoeDepth \cite{bhat2023zoedepth} to train a depth decoder and metric bins module on top of the Depth Anything \cite{yang2024depthany} pre-trained encoder, in order to obtain absolute depth estimations. We train the system for aerial environments using only sparse labels generated for both China and Germany scenes. 
As this model directly produces absolute estimations, no post-inference scaling is required, but it is included in our study as an alternate option to TanDepth for absolute depth systems. The default settings are used for training, which is carried out on a set of 3400 frames for 20 epochs.

We compute and report the widely adopted error metrics of Absolute Relative Error (AbsRel), Square Relative Error (SqRel), Root Mean Square Error (RMSE) and Root Mean Square Log Error (LogRMSE), as well as the Threshold Accuracy metrics $\delta<1.25^T$, where $T\in{1,2,3}$ and $\delta=\mathrm{max}(D^*/D,D/D^*)$. 
It can be observed in Table \ref{tab:mainresults} that this metric is close to saturation in the aerial domain, as $\delta^1$ allows, for example, errors of up to \qty{12.5}{\metre} $(25\%)$ for points at \qty{50}{\metre}. To improve relevance, we propose a tenfold decrease of the threshold: $\Bar{\delta}<1.025^T$.  

\begin{figure*}
    \scriptsize
    \centering
	\includegraphics[width=18cm]{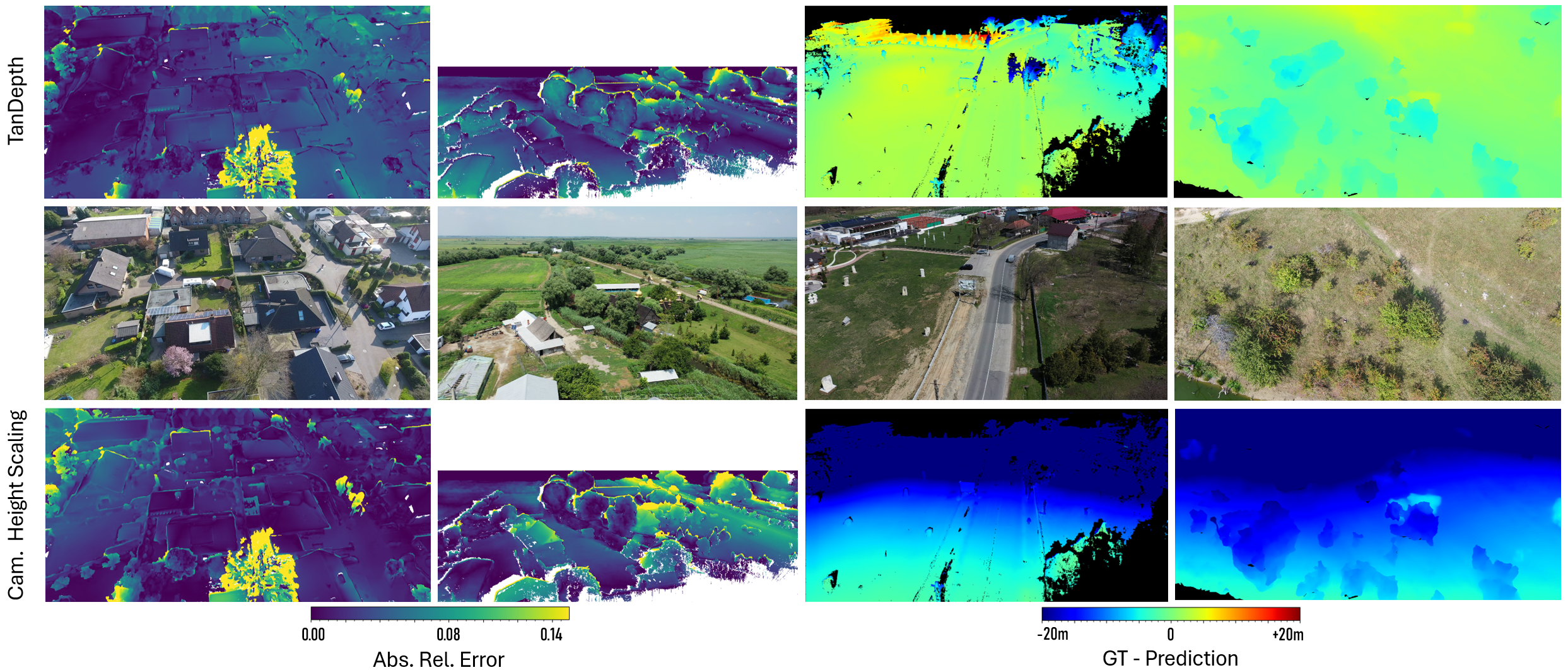}
	\caption{TanDepth Qualitative Results. First two columns: absolute relative error maps for frames in UAVid and UFO Depth Chilia. Last two columns: Error maps depicting signed difference between reference depth and scaled predictions, in UFO Depth Oveselu and WildUAV. Top row features predictions scaled with TanDepth, while the bottom row shows scaling using method based on camera height.}
	\label{fig:qual}
\end{figure*}

\subsection{Absolute Depth Estimation using TanDepth Scaling} \label{sect:main_results}

Table \ref{tab:mainresults} reports results for the proposed TanDepth scaling method in comparison with alternate strategies for generating metric depth results starting from a relative MDE. 
Tested scaling methods include using fixed scaling factors precomputed on a validation set (in our case, on the UAVid Germany Val set), using camera height-based scaling adapted for UAVs (Cam. Height), the trained ZoeDepth metric model and the baseline, evaluation-only strategy of scaling each frame with its corresponding Reference depth map. 

The quantitative evaluation highlights significant improvements the proposed scaling methods consistently brings, across all tested datasets, building towards closing the gap between inference-time methods and the offline baseline, Reference scaling. 
Fixed factor scaling is typically the worst performer, as one would expect, as the network's outputs skew away from the distribution observed on the validation set used for computing the scalars.
The performance of camera height-based scaling significantly drops, as predicted, for environments featuring uneven, hilly terrain, such as UFO Depth Oveselu and WildUAV, compared to the other, flat areas.
Learning to directly infer metric depth, although specifically trained on aerial domain data (UAVid) and starting from the same powerful encoder (Depth Anything \cite{yang2024depthany}), shows its generalization limitations, through poor results on out-of-distribution data (UFO Depth, WildUAV). This comes in line with challenges observed for the UniDepth \cite{piccinelli2024unidepth} metric MDE, tested in Section \ref{sect:ablationstudies} which performs poorly outside its target automotive domain. Moreover, the performance degradation on out-of-distribution evaluations we observe for ZoeDepth in the aerial domain is comparable to that observed by others in the automotive context \cite{yang2024depthany, bhat2023zoedepth}. 

Qualitative results using TanDepth and camera height-based scaling are illustrated in Fig. \ref{fig:qual}. These confirm that the proposed method handles sloped scenes (last two columns) significantly better than the height based approach, which tends to flatten elevations further downfield (predicted value is greater compared to true depth). For flatter scenes, the differences between the two are comparable. Another aspect to note are errors around large vegetation, appearing due to differences in how the depth model and the multi-view-stereo method used to generate the reference reconstruct such objects.

To validate TanDepth's applicability in live operation systems, we measure the average execution times for the proposed components. 
The test was conducted using the official Python wrapper for the CSF algorithm which lacks advanced optimizations such as parallelism.
Evaluation was carried out on the UFO Depth \cite{licuaret2022ufo} Oveselu scene, using a \qty{4.8}{\square\kilo\metre} densified GDEM tile containing 221000 points (requiring 5300 KB of memory) -- the original TanDEM-X data for the area consists of 6900 points (167 KB). Projecting the GDEM points took, on average, \qty{2.5}{\milli\second} and the occlusion handling pass took \qty{3.5}{\milli\second} when computed on the GPU. On an Intel Core i7-6700K CPU, the Cloth Simulation Filter execution adds another \qty{253}{\milli\second}, which can be further optimized down to \qty{90}{\milli\second} with minimal loss in quality, as presented in Section \ref{sect:discuss}.

\textbf{Comparison with Optical Flow-based approaches.} We implement an alternative approach for computing metric depth maps based on depth triangulation using correspondences between optical flow displacements and (metric) relative poses, operating on frame pairs. We follow the depth triangulation method presented in \cite{guizilini2022learning}, using the pretrained RAFT \cite{teed2020raft} optical flow network along with the same pose data used in TanDepth experiments. Table \ref{tab:oflow} presents results of this experiment on the UFO Depth Oveselu scene: while the raw triangulated depth shows good threshold accuracy, its performance across other metrics is significantly worse. Using it in the scaling process instead of the GDEM-based metric ground map $\tilde{g}^*$ yields good performance, but lower than that of the proposed TanDepth method. The utility of the proposed CSF-based ground segmentation module is also confirmed in this new experimental setting, its use improving the scaling performance for the triangulated depth.

It must be noted that using methods relying on optical flow effectively hinders the ability of the system to perform \textit{single image} depth estimation, as a frame pair is required. Moreover, integrating such methods increases the runtime of the entire system - the optical flow inference alone takes, on average, \qty{101}{\milli\second} (with the default 20 RAFT iterations), with additional computation for triangulation. Even when forgoing the use of the CSF-based ground segmentation for both approaches, TanDepth outperforms scaling using triangulated depth maps, with significantly lower computation overhead.

\begin{table}
\centering
\scriptsize
\caption{Comparison with Optical Flow approaches. Results on the UFO Depth Oveselu scene for the depth triangulation approach based on Optical Flow \cite{guizilini2022learning}: raw performance of the triangulated depth (first row), followed by its use as a metric map for scaling Depth Anything (DA) -- with and without CSF-based ground segmentation, compared to TanDepth scaling (last row).}
\begin{tabular}{ccccccc}
\toprule
Depth Type                                                           & AbsRel$\downarrow$ & SqRel$\downarrow$ & RMSE$\downarrow$ & $\Bar{\delta}_1\uparrow$ & $\Bar{\delta}_2\uparrow$ & $\Bar{\delta}_3\uparrow$ \\ \hline
\begin{tabular}[c]{@{}c@{}}Raw Triangulation\\OFlow (TOF)\end{tabular} & 0.054              & 1.929             & 9.087            & 0.461                    & 0.579                    & 0.641                    \\
\begin{tabular}[c]{@{}c@{}}Scaled DA:\\TOF w/o CSF\end{tabular}       & 0.060              & 0.656             & 6.680            & 0.348                    & 0.596                    & 0.746                    \\
\begin{tabular}[c]{@{}c@{}}Scaled DA:\\TOF w/ CSF\end{tabular}        & 0.051              & 0.485             & 6.095            & 0.387                    & 0.634                    & 0.780                    \\
\begin{tabular}[c]{@{}c@{}}Scaled DA:\\TanDepth\end{tabular}                                                          & 0.048              & 0.344             & 5.498            & 0.330                    & 0.592                    & 0.784 \\ \bottomrule                 
\end{tabular}
\footnotesize
\label{tab:oflow}
\end{table}

\begin{table}
\centering
\scriptsize
\caption{Ground Segmentation Ablation Study. Masking out non-ground patches with the Cloth Simulation Filter-based segmentation improves TanDepth performance. On a subset of frames in UAVid Germany for which semantic annotations are provided, the performance of the two segmentation types is similar.}
\begin{tabular}{cccccccc}
\toprule
Scene                    & Filter & AbsRel$\downarrow$ & SqRel$\downarrow$ & RMSE$\downarrow$  & $\Bar{\delta}_1\uparrow$  & $\Bar{\delta}_2\uparrow$  & $\Bar{\delta}_3\uparrow$   \\ \hline
\multirow{3}{*}{\begin{tabular}[c]{@{}c@{}}UAVid \\ Germany \\ (Semantic)\end{tabular}} & None      & 0.044                & 0.250               & 3.648            & 0.365          & 0.756          & 0.890          \\
                                                                                        & Sem. & 0.030                & 0.142               & 2.801            & 0.577          & 0.867          & 0.942          \\
                                                                                        & CSF       & 0.031                & 0.144               & 2.809            & 0.542          & 0.869          & 0.943          \\ \noalign{\vskip -5pt} \\
\multirow{2}{*}{\begin{tabular}[c]{@{}c@{}}UFO Depth \\ Oveselu\end{tabular}}           & None      & 0.055                & 0.405               & 5.788            & 0.291          & 0.543          & 0.733          \\
                                                                                        & CSF       & 0.048                & 0.344               & 5.498            & 0.330          & 0.592          & 0.784          \\ \noalign{\vskip -5pt} \\
\multirow{2}{*}{\begin{tabular}[c]{@{}c@{}}UFO Depth \\ Chilia\end{tabular}}            & None      & 0.075                & 1.227               & 10.617           & 0.276          & 0.475          & 0.625          \\
                                                                                        & CSF       & 0.040                & 0.442               & 7.520            & 0.427          & 0.715          & 0.872          \\ 
                                                                                       \bottomrule

\end{tabular}
\footnotesize
\label{tab:ablfiltering}
\end{table}

\begin{table*}
\centering
\scriptsize
\caption{MDE Model Comparison. For each metric, the best result is in bold, per scene. UniDepth is an absolute (metric) depth model, yet its performance on aerial scenes is quite poor prior to applying a scaling approach.}
\begin{tabular}{crrcccccc}
\toprule
Scene                                                                   & Model                & Scaling                      & AbsRel$\downarrow$ & SqRel$\downarrow$ & RMSE$\downarrow$  & $\Bar{\delta}_1\uparrow$  & $\Bar{\delta}_2\uparrow$  & $\Bar{\delta}_3\uparrow$        \\ \hline
\multirow{7}{*}{\begin{tabular}[c]{@{}c@{}}UFOD\\ Oveselu\end{tabular}} & \multirow{2}{*}{Depth Anything}  & \textit{Reference}                  & {\textbf{\textit{0.040}}} & \textit{0.285}       & {\textbf{\textit{5.057}}} & {\textbf{\textit{0.430}}} & \textit{0.701}       & {\textbf{\textit{0.849}}} \\
                                                                        &                      & TanDepth                     & 0.048                & 0.344                & 5.498                & 0.330                & 0.592                & 0.784                \\ \noalign{\vskip -6pt} \\ 
                                                                        & \multirow{2}{*}{Depth Anything V2} & \textit{Reference}                  & 0.042                & 0.311                & 5.252                & 0.426                & 0.682                & 0.829                \\
                                                                        &                      & TanDepth                     & 0.051                & 0.380                & 5.726                & 0.309                & 0.567                & 0.757                \\ \noalign{\vskip -6pt} \\  
                                                                        & \multirow{3}{*}{UniDepth} & \textit{None*}               & 0.581                & 32.957               & 56.982               & 0.000                & 0.000                & 0.001                \\
                                                                        &                      & \textit{Reference}                  & {\textbf{\textit{0.040}}} & {\textbf{\textit{0.282}}} & \textit{5.180}       & \textit{0.424}       & {\textbf{\textit{0.703}}} & \textit{0.847}       \\
                                                                        &                      & TanDepth                     & 0.047                & 0.310                & 5.233                & 0.326                & 0.607                & 0.797                \\ \noalign{\vskip -4pt} \\
\multirow{7}{*}{\begin{tabular}[c]{@{}c@{}}UFOD\\ Chilia\end{tabular}}  & \multirow{2}{*}{Depth Anything}  & \textit{Reference}                  & \textit{0.035}       & \textit{0.362}       & \textit{7.064}       & \textit{0.496}       & \textit{0.779}       & \textit{0.904}       \\
                                                                        &                      & TanDepth                     & 0.040                & 0.442                & 7.520                & 0.427                & 0.715                & 0.872                \\ \noalign{\vskip -6pt} \\ 
                                                                        & \multirow{2}{*}{Depth Anything V2} & \textit{Reference}                  & {\textbf{\textit{0.033}}} & {\textbf{\textit{0.328}}} & {\textbf{\textit{6.661}}} & {\textbf{\textit{0.521}}} & {\textbf{\textit{0.811}}} & {\textbf{\textit{0.922}}} \\ 
                                                                        &                      & TanDepth                     & 0.039                & 0.410                & 7.206                & 0.424                & 0.736                & 0.888                \\ \noalign{\vskip -6pt} \\ 
                                                                        & \multirow{3}{*}{UniDepth} & \textit{None*}               & 0.513                & 43.361               & 84.116               & 0.001                & 0.002                & 0.004                \\
                                                                        &                      & \textit{Reference}                  & \textit{0.040}       & \textit{0.440}       & \textit{7.905}       & \textit{0.440}       & \textit{0.716}       & \textit{0.859}       \\
                                                                        &                      & \multicolumn{1}{l}{TanDepth} & 0.046                & 0.517                & 8.122                & 0.386                & 0.653                & 0.815                \\ \noalign{\vskip -4pt} \\
\multicolumn{1}{l}{\multirow{7}{*}{WildUAV}}                            & \multirow{2}{*}{Depth Anything}  & \textit{Reference}                  & {\textbf{\textit{0.041}}} & {\textbf{\textit{0.157}}} & {\textbf{\textit{2.560}}} & {\textbf{\textit{0.448}}} & {\textbf{\textit{0.708}}} & {\textbf{\textit{0.846}}} \\
\multicolumn{1}{l}{}                                                    &                      & TanDepth                     & 0.072                & 1.269                & 4.106                & 0.322                & 0.531                & 0.670                \\ \noalign{\vskip -6pt} \\ 
\multicolumn{1}{l}{}                                                    & \multirow{2}{*}{Depth Anything V2} & \textit{Reference}                  & \textit{0.043}       & \textit{0.180}       & \textit{2.763}       & \textit{0.427}       & \textit{0.686}       & \textit{0.830}       \\
\multicolumn{1}{l}{}                                                    &                      & TanDepth                     & 0.106                & 2.510                & 5.723                & 0.254                & 0.431                & 0.552                \\ \noalign{\vskip -6pt} \\ 
\multicolumn{1}{l}{}                                                    & \multirow{3}{*}{UniDepth} & \textit{None*}               & \textit{0.353}       & \textit{6.908}       & \textit{18.958}      & \textit{0.000}       & \textit{0.001}       & \textit{0.002}       \\
\multicolumn{1}{l}{}                                                    &                      & \textit{Reference}                  & \textit{0.048}       & \textit{0.240}       & \textit{3.396}       & \textit{0.351}       & \textit{0.625}       & \textit{0.798}       \\
\multicolumn{1}{l}{}                                                    &                      & \multicolumn{1}{l}{TanDepth} & 0.159                & 3.860                & 12.016               & 0.175                & 0.305                & 0.404            \\ 
\bottomrule   
\end{tabular}
\label{tab:modelsbenchmark}
\end{table*}

\subsection{TanDepth Ablation Studies} \label{sect:ablationstudies}

\textbf{Effect of Ground Segmentation.} We perform experiments to measure the effect ground segmentations have on the quality of the scaled estimations. First, we evaluate the proposed CSF-based filtering which can be applied directly on the initial (unscaled) depth, comparing it with evaluations in which no ground filtering is applied, over three scenes. In addition, on UAVid Germany, we use available semantic annotations to compare how a semantic model-based segmentation would perform for the filtering task. To obtain a similar ground segmentation from the semantic image, we select pixels classified as either "Road" or "Low Vegetation", the two UAVid classes which encompass the majority of ground points. Note that for this test, we use only frames for which semantic annotations are available, which would be used to train the segmentation model and represent an ideal segmentation.

The results presented in Table \ref{tab:ablfiltering} confirm that limiting the scaling process to using only ground patches in the observed camera frame leads to improvements in the resulting depth estimation. At the same time, using the semantic annotations leads to similar scaling performance, despite relying on more complex data.

\textbf{MDE Model Comparison.} Table \ref{tab:modelsbenchmark} presents results comparing the performance of three state of the art MDE models in the aerial context. Each model is evaluated both using Reference scaling, as well as in conjunction with the proposed TanDepth approach. 
As a metric MDE model, we additionally evaluate UniDepth \cite{piccinelli2024unidepth} without any type of scaling running it with provided camera intrinsics.

The two versions of Depth Anything behave similarly, with the first version outperforming the second one in two of the tested scenes. Without scaling, UniDepth's excellent zero-shot performance in the automotive domain, for which it was designed, does not translate well to aerial viewpoints. 
However, by evaluating it in conjunction with an additional scaling mechanism, its performance dramatically ameliorates, supporting the view that limitations in MDE generalization are strongly linked with the issue of scale.  

\begin{figure}
    \centering
	\includegraphics[width=9cm]{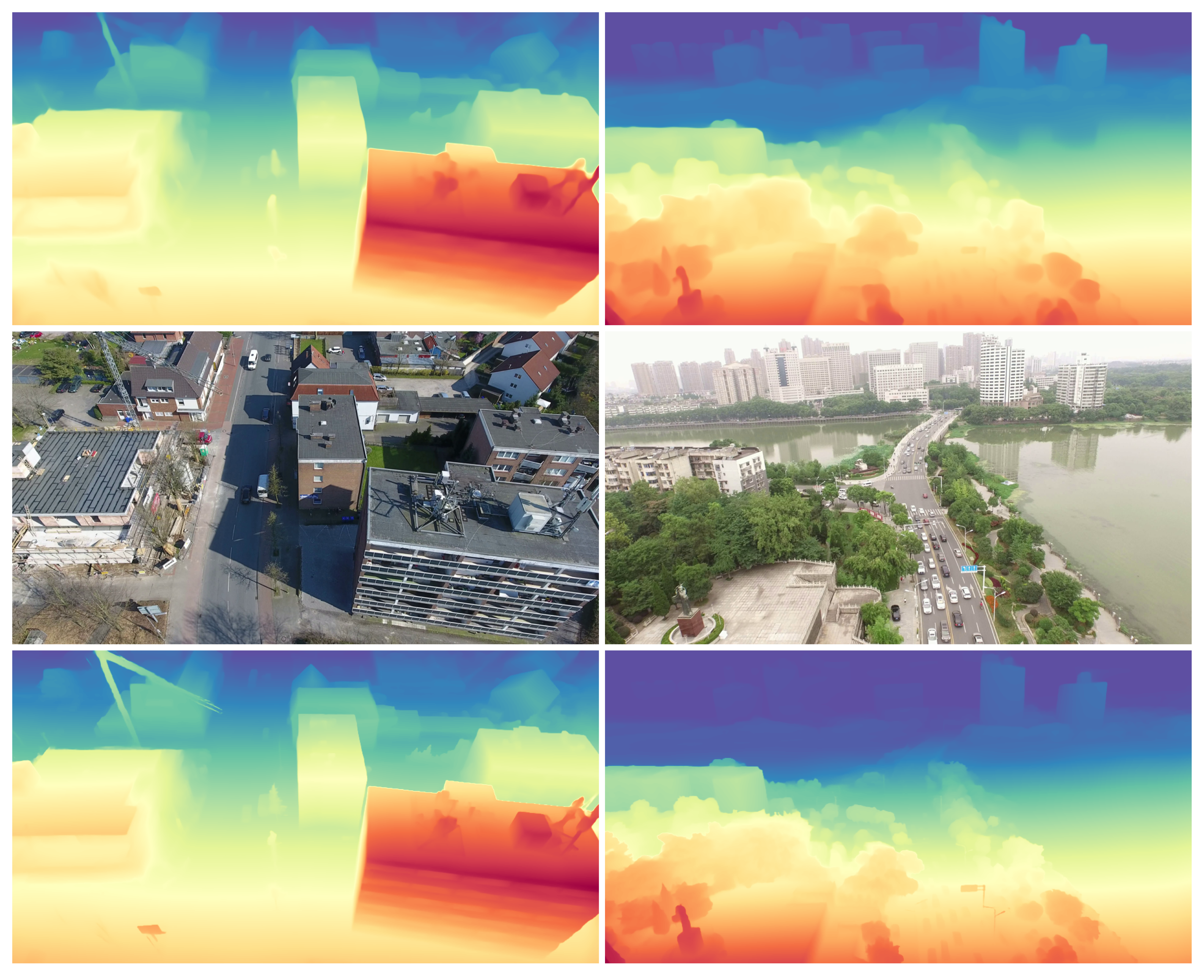}
	\caption{Qualitative comparison of disparity maps generated by Depth Anything V1 (1st row) and V2 (3rd row) on aerial data. V2's training setup clearly improves fine detail performance and also solves issues with occasional artefacts on water reflections (2nd row, right-side buildings).}
	\label{fig:davda2}
\end{figure}

Depth Anything V2 \cite{yang2024depthanyv2}, as the successor of Depth Anything \cite{yang2024depthany}, inherits many of its advantages while better preserving fine details. 
This can be clearly observed in the qualitative comparison between the two, in Fig. \ref{fig:davda2}, such as for the construction crane in the first image. 
Of interest for aerial applications is also V2's improved handling of reflective water surface, such as those in the second image (below top right skyscrapers). 
The level at which the model is able to resolve fine details at significant distances is impressive, and highlights the limitations of standard quantitative evaluation metrics which don't capture such differences.

\textbf{Effect of GDEM Densification.} A separate experiment, presented in Table \ref{tab:ablsamp} was conducted on the UFO Depth Oveselu scene to evaluate the impact of GDEM data density has on the scaling performance. The default density value of \qty[per-mode = symbol]{0.05}{\points\per\square\metre} used in all of the presented experiments in the paper is compared with two other values, as well as to using the original GDEM data directly. While the performance deltas are limited in magnitude, increasing the density of the data is shown to improve the scaling performance. Of note is that higher densities also reduce the number of frames for which scaling can't take place due to lack of projected ground GDEM points. At the same time, increasing the GDEM density inherently increases the number of projections and thus, computational complexity.

\begin{table}
\scriptsize
\centering
\caption{Densification Ablation. Comparing the effects of different GDEM Densification strengths on UFO Depth Oveselu scene. The 0.05 $\mathrm{pts/m}^2$ value is the default used throughout the other experiments in the paper.}
\begin{tabular}{lcccccc}
\toprule
Points/$\mathrm{m}^2$   & AbsRel$\downarrow$ & SqRel$\downarrow$ & RMSE$\downarrow$  & $\Bar{\delta}_1\uparrow$  & $\Bar{\delta}_2\uparrow$  & $\Bar{\delta}_3\uparrow$  \\ \hline
Original          & 0.051    & 0.374   & 5.676      & 0.323 & 0.581 & 0.765 \\
0.005         & 0.049    & 0.354   & 5.561      & 0.326 & 0.586 & 0.776 \\
\textbf{0.05} & 0.048    & 0.344   & 5.498      & 0.330 & 0.592 & 0.784 \\
0.5 & 0.049    & 0.352   & 5.559      & 0.329 & 0.593 & 0.783 \\
\bottomrule
\end{tabular}
\footnotesize
\label{tab:ablsamp}
\end{table}

\section{Discussion} \label{sect:discuss}

\textbf{Result Analysis.} The main experimental results confirm that the proposed TanDepth method represents a significant leap forward in performance and generalization towards the key capability of metric scale recovery for aerial monocular depth estimation. This enables the use of relative depth estimators, such as recent foundation models and self-supervised MDEs, in a variety of applications that require metric scene geometry reconstruction. Our proposal outperforms alternative methods across all tested scenes, which sample a diverse set of conditions. Given a particular relative depth model, a significant milestone of any scale recovery method is to match the performance of using Reference scaling, the standard procedure used in offline evaluations. Analyzing the results through this perspective, TanDepth comes closest to matching Reference scaling in UAVid Germany, the most structured scene in the test set, captured at a $45 ^\circ$ pitch angle, above flat terrain. Very good results are also obtained over the UFO Depth scenes: Oveselu, featuring hilly terrain and a mix of man-made and natural areas, and Chilia, a flat scene with even fewer structured elements. The higher overall error rates in the latter are motivated, in part, due to the larger and deeper extent of the evaluated area (\qtyrange{50}{250}{\metre}) as well as the shallower pitch angle.

While still outperforming other methods, the difference in performance between TanDepth and Reference scaling is largest in the WildUAV scene, defined by few man-made structures, presence of a small lake, significant terrain variations visible in individual frames, along with a slightly lower flight trajectory compared to the other scenes. Error metrics are, in part, elevated due to a large single-frame error caused by a poor ground segmentation -- omitting that frame from evaluation improves the Absolute Relative, RMS and $\Bar{\delta_1}$ errors to 0.063, 3.729 and 0.323 respectively. Camera Height based scaling ranks second best after TanDepth, with relatively good performance in flat scenes but significant degradation in Oveselu and WildUAV, caused, as expected, by uneven terrain. The Fixed factor and ZoeDepth scaling solutions fail to generalize beyond the UAVid Germany on which they were tuned and trained, respectively.

\textbf{Method Parameters.} The proposed system features few parameters that need tuning during operation. Parameters pertaining to the occlusion handling process should only be modified (smaller window, larger depth-aware threshold) if the stability of the system is affected by large numbers of rejected points. In practical applications, the user is suggested to adjust the parameters of the Cloth Simulation Filter-based \cite{zhang2016csf} ground segmentation in order to improve its output for the specific terrain profile. Running the CSF algorithm on unscaled point clouds generated by the relative depth model could affect the values of the parameters. By design, our adaptations handle this type of variability, allowing the end user to directly follow recommendations \cite{zhang2016csf} and procedural knowledge for the widely used original CSF implementation: higher rigidity values should be used when terrain is known flat (we use a low value of 1), lower values for the cloth resolution can produce better results yet increase processing times and should be positively correlated with changes to the class threshold.

For performing the rough initial scaling in the adapted CSF, the $\Bar{s}$ and $\Bar{t}$ parameters need to be precomputed as fixed median scaling parameters on a validation set -- as these are model dependent, they must be computed ahead of time (once) for each type of relative depth model intended for pairing with TanDepth. For the GDEM densification process, the ablation study presented in Table \ref{tab:ablsamp} motivates our choice for the sampling value of \qty[per-mode = symbol]{0.05}{\points\per\square\metre}. While the densification process' main role (avoiding cases with few or no valid GDEM points visible in the frame) is ensured in most cases using this value, the sampling parameter should be increased when performing flights at significantly lower altitudes and could be decreased in the case of higher altitudes (reducing computational cost).

\textbf{CSF Speed-up.} As presented in Section \ref{sect:main_results}, the CSF segmentation is the largest contributor to the overall execution time. As such, we set out to investigate ways to speed-up this step with minimal degradation of the results. Intuitively, increasing the value of the cloth resolution CSF parameter, and thus using a coarser mesh brings significant improvements in the execution time. This, however, can affect the stability of the segmentation and requires frequent manual readjustments of the class threshold parameter, leading to an overall less mature system.

We devise an alternate approach, which instead lowers the resolution of the depth image used as input for the CSF, decreasing, in effect, the number of 3D points used to represent the scene and thus, the number of interactions between it and the simulated cloth. Testing on the UFO Depth \cite{licuaret2022ufo} Oveselu scene validates that by keeping the original CSF parameters and using a $64 \times 128$px input lowers the execution time from \qty{253}{\milli\second} down to \qty{90}{\milli\second}, with increases of only $0.004$ and $0.035$  in Squared Relative and Root Mean Square error metrics, respectively. 
Additional gains in performance could be achieved by transitioning to a GPU implementation and optimizing for on-board compute platforms such as NVIDIA Jetson TX2. These represent engineering challenges which are beyond the scope of the current paper.

\textbf{Future developments.} Additional improvements for TanDepth could be investigated and developed in order to further increase the ease of use of the method, on the path towards on-board integration. A deployment-ready version of the system would benefit by adding a more advanced solution for storing densified GDEM data for wide areas, by dividing it in tiles that are stored in a data structure such as a quadtrees. This would enable fast retrieval of nearby tiles based on current position and heading, minimizing location constraints on the surveyed area.

A second development path would aim to automate the selection of the CSF ground segmentation parameters. The principal factors guiding the adjustments are the flight altitude and camera pitch, along with the terrain profile. The first two effectively determine the Ground Sampling Distance during the flight, or a range of values, depending on the position in the image and camera pitch. A pre-flight processing step could also classify the degree of terrain variations beneath the flight area, based on the GDEM data.

Considering TanDepth's design as a post-processing step that can be applied to a wide variety of relative depth models, it would be of scientific value to develop and evaluate refinements for use in systems based on different image domains such as infrared/thermal by pairing with an appropriate depth model \cite{shin2023ir}. Such an ensemble could enable further downstream tasks such as 3D localization for object detection \cite{zhang2021deep}. Additionally, the scaled metric depth maps resulting from the proposed system could be used to augment further classification tasks for environmental conditions/places using modern methods \cite{zhang2024part, zhang2024cognition}.

\section{Conclusion}
In this work, we present TanDepth, a practical inference-time scale-recovery approach for monocular depth estimation that leverages the recently released TanDEM-X public Global Digital Elevation Model. Projected GDEM points represent scale depth anchors in the image frame that are then used to scale relative estimations, including those coming from scale- and shift-invariant models. 
Our adaptation for the Cloth Simulation Filter enables its use for segmenting ground pixels based on relative depth estimations and thus allows the scaling mechanism to be constrained to use only ground points. 

The proposed system, when paired with modern relative MDE models, is shown to provide high-quality metric depth maps with good generalization performance across a variety of outdoor UAV scenes, including from the new UAVid-3D-Scenes dataset we develop for public release. TanDepth outperforms other types of inference-time scale recovery methods, including the camera height-based method which we adapt to the aerial domain.

\section*{Acknowledgment}
The OpenAI ChatGPT-4o LLM was used for grammar enhancement of selected phrases and for creating graphical elements in Fig. 1.

\bibliographystyle{IEEEtran}
\bibliography{bib} 

\begin{thebibliography}{10}
\providecommand{\url}[1]{#1}
\csname url@samestyle\endcsname
\providecommand{\newblock}{\relax}
\providecommand{\bibinfo}[2]{#2}
\providecommand{\BIBentrySTDinterwordspacing}{\spaceskip=0pt\relax}
\providecommand{\BIBentryALTinterwordstretchfactor}{4}
\providecommand{\BIBentryALTinterwordspacing}{\spaceskip=\fontdimen2\font plus
\BIBentryALTinterwordstretchfactor\fontdimen3\font minus \fontdimen4\font\relax}
\providecommand{\BIBforeignlanguage}[2]{{%
\expandafter\ifx\csname l@#1\endcsname\relax
\typeout{** WARNING: IEEEtran.bst: No hyphenation pattern has been}%
\typeout{** loaded for the language `#1'. Using the pattern for}%
\typeout{** the default language instead.}%
\else
\language=\csname l@#1\endcsname
\fi
#2}}
\providecommand{\BIBdecl}{\relax}
\BIBdecl

\bibitem{nex2022uav}
F.~Nex, C.~Armenakis, M.~Cramer, D.~A. Cucci, M.~Gerke, E.~Honkavaara, A.~Kukko, C.~Persello, and J.~Skaloud, ``{UAV} in the advent of the twenties: Where we stand and what is next,'' \emph{ISPRS journal of photogrammetry and remote sensing}, vol. 184, pp. 215--242, 2022.

\bibitem{akhloufi2021appfire}
M.~A. Akhloufi, A.~Couturier, and N.~A. Castro, ``Unmanned aerial vehicles for wildland fires: Sensing, perception, cooperation and assistance,'' \emph{Drones}, vol.~5, no.~1, p.~15, 2021.

\bibitem{lyu2023appsar}
M.~Lyu, Y.~Zhao, C.~Huang, and H.~Huang, ``Unmanned aerial vehicles for search and rescue: A survey,'' \emph{Remote Sensing}, vol.~15, no.~13, p. 3266, 2023.

\bibitem{mildenhall2021nerf}
B.~Mildenhall, P.~P. Srinivasan, M.~Tancik, J.~T. Barron, R.~Ramamoorthi, and R.~Ng, ``Nerf: Representing scenes as neural radiance fields for view synthesis,'' \emph{Communications of the ACM}, vol.~65, no.~1, pp. 99--106, 2021.

\bibitem{ranftl2020midas}
R.~Ranftl, K.~Lasinger, D.~Hafner, K.~Schindler, and V.~Koltun, ``Towards robust monocular depth estimation: Mixing datasets for zero-shot cross-dataset transfer,'' \emph{IEEE transactions on pattern analysis and machine intelligence}, vol.~44, no.~3, pp. 1623--1637, 2020.

\bibitem{yang2024depthany}
L.~Yang, B.~Kang, Z.~Huang, X.~Xu, J.~Feng, and H.~Zhao, ``Depth anything: Unleashing the power of large-scale unlabeled data,'' in \emph{2024 IEEE/CVF Conference on Computer Vision and Pattern Recognition (CVPR)}, 2024, pp. 10\,371--10\,381.

\bibitem{yang2024depthanyv2}
\BIBentryALTinterwordspacing
L.~Yang, B.~Kang, Z.~Huang, Z.~Zhao, X.~Xu, J.~Feng, and H.~Zhao, ``Depth anything v2,'' in \emph{The Thirty-eighth Annual Conference on Neural Information Processing Systems}, 2024. [Online]. Available: \url{https://openreview.net/forum?id=cFTi3gLJ1X}
\BIBentrySTDinterwordspacing

\bibitem{florea2022survey}
H.~Florea and S.~Nedevschi, ``Survey on monocular depth estimation for unmanned aerial vehicles using deep learning,'' in \emph{2022 IEEE 18th International Conference on Intelligent Computer Communication and Processing (ICCP)}.\hskip 1em plus 0.5em minus 0.4em\relax IEEE, 2022, pp. 319--326.

\bibitem{zhousfml}
T.~Zhou, M.~Brown, N.~Snavely, and D.~G. Lowe, ``Unsupervised learning of depth and ego-motion from video,'' in \emph{Proceedings of the IEEE conference on computer vision and pattern recognition}, 2017, pp. 1851--1858.

\bibitem{rizzoli2017tdem}
P.~Rizzoli, M.~Martone, C.~Gonzalez, C.~Wecklich, D.~B. Tridon, B.~Br{\"a}utigam, M.~Bachmann, D.~Schulze, T.~Fritz, M.~Huber \emph{et~al.}, ``Generation and performance assessment of the global {TanDEM-X} digital elevation model,'' \emph{ISPRS Journal of Photogrammetry and Remote Sensing}, vol. 132, pp. 119--139, 2017.

\bibitem{zhang2016csf}
W.~Zhang, J.~Qi, P.~Wan, H.~Wang, D.~Xie, X.~Wang, and G.~Yan, ``An easy-to-use airborne {LiDAR} data filtering method based on cloth simulation,'' \emph{Remote sensing}, vol.~8, no.~6, p. 501, 2016.

\bibitem{lyu2020uavid}
Y.~Lyu, G.~Vosselman, G.-S. Xia, A.~Yilmaz, and M.~Y. Yang, ``{UAVid}: A semantic segmentation dataset for {UAV} imagery,'' \emph{ISPRS journal of photogrammetry and remote sensing}, vol. 165, pp. 108--119, 2020.

\bibitem{licuaret2022ufo}
V.~Lic{\u{a}}ret, V.~Robu, A.~Marcu, D.~Costea, E.~Slu{\c{s}}anschi, R.~Sukthankar, and M.~Leordeanu, ``Ufo depth: Unsupervised learning with flow-based odometry optimization for metric depth estimation,'' in \emph{2022 International Conference on Robotics and Automation (ICRA)}.\hskip 1em plus 0.5em minus 0.4em\relax IEEE, 2022, pp. 6526--6532.

\bibitem{florea2021wilduav}
H.~Florea, V.-C. Miclea, and S.~Nedevschi, ``{WildUAV}: Monocular {UAV} dataset for depth estimation tasks,'' in \emph{2021 IEEE 17th International Conference on Intelligent Computer Communication and Processing (ICCP)}.\hskip 1em plus 0.5em minus 0.4em\relax IEEE, 2021, pp. 291--298.

\bibitem{ranftl2021dpt}
R.~Ranftl, A.~Bochkovskiy, and V.~Koltun, ``Vision transformers for dense prediction,'' in \emph{Proceedings of the IEEE/CVF international conference on computer vision}, 2021, pp. 12\,179--12\,188.

\bibitem{guizilini2023zerodepth}
V.~Guizilini, I.~Vasiljevic, D.~Chen, R.~Ambruș, and A.~Gaidon, ``Towards zero-shot scale-aware monocular depth estimation,'' in \emph{Proceedings of the IEEE/CVF International Conference on Computer Vision}, 2023, pp. 9233--9243.

\bibitem{piccinelli2024unidepth}
L.~Piccinelli, Y.-H. Yang, C.~Sakaridis, M.~Segu, S.~Li, L.~V. Gool, and F.~Yu, ``{UniDepth}: Universal monocular metric depth estimation,'' in \emph{2024 IEEE/CVF Conference on Computer Vision and Pattern Recognition (CVPR)}, 2024, pp. 10\,106--10\,116.

\bibitem{fu2018deep}
H.~Fu, M.~Gong, C.~Wang, K.~Batmanghelich, and D.~Tao, ``Deep ordinal regression network for monocular depth estimation,'' in \emph{Proceedings of the IEEE conference on computer vision and pattern recognition}, 2018, pp. 2002--2011.

\bibitem{bhat2021adabins}
S.~F. Bhat, I.~Alhashim, and P.~Wonka, ``Adabins: Depth estimation using adaptive bins,'' in \emph{Proceedings of the IEEE/CVF conference on computer vision and pattern recognition}, 2021, pp. 4009--4018.

\bibitem{bhat2022localbins}
S.~F. Bhat, I.~Alhashim, and P.~Wonka, ``Localbins: Improving depth estimation by learning local distributions,'' in \emph{European Conference on Computer Vision}.\hskip 1em plus 0.5em minus 0.4em\relax Springer, 2022, pp. 480--496.

\bibitem{bhat2023zoedepth}
S.~F. Bhat, R.~Birkl, D.~Wofk, P.~Wonka, and M.~M{\"u}ller, ``Zoedepth: Zero-shot transfer by combining relative and metric depth,'' \emph{arXiv preprint arXiv:2302.12288}, 2023.

\bibitem{miclea2021monocular}
V.-C. Miclea and S.~Nedevschi, ``Monocular depth estimation with improved long-range accuracy for {UAV} environment perception,'' \emph{IEEE Transactions on Geoscience and Remote Sensing}, vol.~60, pp. 1--15, 2021.

\bibitem{miclea2023dynamic}
V.-C. Miclea and S.~Nedevschi, ``Dynamic semantically guided monocular depth estimation for uav environment perception,'' \emph{IEEE Transactions on Geoscience and Remote Sensing}, 2023.

\bibitem{madhuanand2020deep}
L.~Madhuanand, F.~Nex, M.~Yang, N.~Paparoditis, C.~Mallet, F.~Lafarge, F.~Remondino, I.~Toschi, T.~Fuse \emph{et~al.}, ``Deep learning for monocular depth estimation from {UAV} images,'' \emph{ISPRS Annals of the Photogrammetry, Remote Sensing and Spatial Information Sciences}, vol.~2, 2020.

\bibitem{fonder2021m4depth}
\BIBentryALTinterwordspacing
M.~Fonder, D.~Ernst, and M.~Van~Droogenbroeck, ``Parallax inference for robust temporal monocular depth estimation in unstructured environments,'' \emph{Sensors}, vol.~22, no.~23, pp. 1--22, November 2022. [Online]. Available: \url{https://doi.org/10.3390/s22239374}
\BIBentrySTDinterwordspacing

\bibitem{gargfirst}
R.~Garg, V.~K. Bg, G.~Carneiro, and I.~Reid, ``Unsupervised cnn for single view depth estimation: Geometry to the rescue,'' in \emph{European conference on computer vision}.\hskip 1em plus 0.5em minus 0.4em\relax Springer, 2016, pp. 740--756.

\bibitem{godardmono2}
C.~Godard, O.~Mac~Aodha, M.~Firman, and G.~J. Brostow, ``Digging into self-supervised monocular depth estimation,'' in \emph{Proceedings of the IEEE/CVF International Conference on Computer Vision}, 2019, pp. 3828--3838.

\bibitem{bian2}
J.-W. Bian, H.~Zhan, N.~Wang, Z.~Li, L.~Zhang, C.~Shen, M.-M. Cheng, and I.~Reid, ``Unsupervised scale-consistent depth learning from video,'' \emph{International Journal of Computer Vision}, pp. 1--17, 2021.

\bibitem{madhuanand2021self}
L.~Madhuanand, F.~Nex, and M.~Y. Yang, ``Self-supervised monocular depth estimation from oblique {UAV} videos,'' \emph{ISPRS journal of photogrammetry and remote sensing}, vol. 176, pp. 1--14, 2021.

\bibitem{hermann2021real}
\BIBentryALTinterwordspacing
M.~Hermann, B.~Ruf, and M.~Weinmann, ``Real-time dense 3d reconstruction from monocular video data captured by low-cost uavs,'' \emph{The International Archives of the Photogrammetry, Remote Sensing and Spatial Information Sciences}, vol. XLIII-B2-2021, pp. 361--368, 2021. [Online]. Available: \url{https://isprs-archives.copernicus.org/articles/XLIII-B2-2021/361/2021/}
\BIBentrySTDinterwordspacing

\bibitem{xue2020toward}
F.~Xue, G.~Zhuo, Z.~Huang, W.~Fu, Z.~Wu, and M.~H. Ang, ``Toward hierarchical self-supervised monocular absolute depth estimation for autonomous driving applications,'' in \emph{2020 IEEE/RSJ International Conference on Intelligent Robots and Systems (IROS)}.\hskip 1em plus 0.5em minus 0.4em\relax IEEE, 2020, pp. 2330--2337.

\bibitem{wagstaffScale}
B.~Wagstaff and J.~Kelly, ``Self-supervised scale recovery for monocular depth and egomotion estimation,'' in \emph{2021 IEEE/RSJ International Conference on Intelligent Robots and Systems (IROS)}.\hskip 1em plus 0.5em minus 0.4em\relax IEEE, 2021, pp. 2620--2627.

\bibitem{li2024radarcam}
H.~Li, Y.~Ma, Y.~Gu, K.~Hu, Y.~Liu, and X.~Zuo, ``Radarcam-depth: Radar-camera fusion for depth estimation with learned metric scale,'' in \emph{2024 IEEE International Conference on Robotics and Automation (ICRA)}, 2024, pp. 10\,665--10\,672.

\bibitem{packnet}
V.~Guizilini, R.~Ambrus, S.~Pillai, A.~Raventos, and A.~Gaidon, ``3d packing for self-supervised monocular depth estimation,'' in \emph{Proceedings of the IEEE/CVF Conference on Computer Vision and Pattern Recognition}, 2020, pp. 2485--2494.

\bibitem{swami2022dwyc}
K.~Swami, A.~Muduli, U.~Gurram, and P.~Bajpai, ``Do what you can, with what you have: Scale-aware and high quality monocular depth estimation without real world labels,'' in \emph{Proceedings of the IEEE/CVF Conference on Computer Vision and Pattern Recognition}, 2022, pp. 988--997.

\bibitem{pirvu2021depth}
M.~Pirvu, V.~Robu, V.~Licaret, D.~Costea, A.~Marcu, E.~Slusanschi, R.~Sukthankar, and M.~Leordeanu, ``Depth distillation: unsupervised metric depth estimation for {UAVs} by finding consensus between kinematics, optical flow and deep learning,'' in \emph{Proceedings of the IEEE/CVF Conference on Computer Vision and Pattern Recognition}, 2021, pp. 3215--3223.

\bibitem{monabench}
\BIBentryALTinterwordspacing
Y.~Pan, B.~Liu, Z.~Liu, H.~Shen, J.~Xu, W.~Fu, and T.~Yang, ``{MoNA Bench}: A benchmark for monocular depth estimation in navigation of autonomous unmanned aircraft system,'' \emph{Drones}, vol.~8, no.~2, 2024. [Online]. Available: \url{https://www.mdpi.com/2504-446X/8/2/66}
\BIBentrySTDinterwordspacing

\bibitem{srtm}
T.~G. Farr and M.~Kobrick, ``{Shuttle Radar Topography Mission} produces a wealth of data,'' \emph{Eos, Transactions American Geophysical Union}, vol.~81, no.~48, pp. 583--585, 2000.

\bibitem{tachikawa2011aster}
T.~Tachikawa, M.~Kaku, A.~Iwasaki, D.~B. Gesch, M.~J. Oimoen, Z.~Zhang, J.~J. Danielson, T.~Krieger, B.~Curtis, J.~Haase \emph{et~al.}, ``{ASTER} global digital elevation model version 2-summary of validation results,'' NASA, Tech. Rep., 2011.

\bibitem{tadono2014alos}
T.~Tadono, H.~Ishida, F.~Oda, S.~Naito, K.~Minakawa, and H.~Iwamoto, ``Precise global {DEM} generation by {ALOS PRISM},'' \emph{ISPRS Annals of the Photogrammetry, Remote Sensing and Spatial Information Sciences}, vol.~2, pp. 71--76, 2014.

\bibitem{shah2018airsim}
S.~Shah, D.~Dey, C.~Lovett, and A.~Kapoor, ``Airsim: High-fidelity visual and physical simulation for autonomous vehicles,'' in \emph{Field and service robotics}.\hskip 1em plus 0.5em minus 0.4em\relax Springer, 2018, pp. 621--635.

\bibitem{fonder2019mid}
M.~Fonder and M.~Van~Droogenbroeck, ``Mid-air: A multi-modal dataset for extremely low altitude drone flights,'' in \emph{Proceedings of the IEEE/CVF conference on computer vision and pattern recognition workshops}, 2019, pp. 0--0.

\bibitem{tartanair2020iros}
W.~Wang, D.~Zhu, X.~Wang, Y.~Hu, Y.~Qiu, C.~Wang, Y.~Hu, A.~Kapoor, and S.~Scherer, ``{TartanAi}r: A dataset to push the limits of visual {SLAM},'' in \emph{2020 IEEE/RSJ International Conference on Intelligent Robots and Systems (IROS)}, 2020.

\bibitem{hermann2024usegeo}
M.~Hermann, M.~Weinmann, F.~Nex, E.~Stathopoulou, F.~Remondino, B.~Jutzi, and B.~Ruf, ``Depth estimation and {3D} reconstruction from {UAV}-borne imagery: Evaluation on the {UseGeo} dataset,'' \emph{ISPRS Open Journal of Photogrammetry and Remote Sensing}, p. 100065, 2024.

\bibitem{bian1}
J.~Bian, Z.~Li, N.~Wang, H.~Zhan, C.~Shen, M.-M. Cheng, and I.~Reid, ``Unsupervised scale-consistent depth and ego-motion learning from monocular video,'' \emph{Advances in neural information processing systems}, vol.~32, pp. 35--45, 2019.

\bibitem{CloudCompare_software}
\BIBentryALTinterwordspacing
{GPL software}, \emph{{CloudCompare} (version 2.13)}, 2024. [Online]. Available: \url{http://www.cloudcompare.org}
\BIBentrySTDinterwordspacing

\bibitem{QGIS_software}
\BIBentryALTinterwordspacing
{QGIS Development Team}, \emph{{QGIS Geographic Information System}}, QGIS Association, 2024. [Online]. Available: \url{https://www.qgis.org}
\BIBentrySTDinterwordspacing

\bibitem{katz2007direct}
S.~Katz, A.~Tal, and R.~Basri, ``Direct visibility of point sets,'' in \emph{ACM SIGGRAPH 2007 papers}, 2007, pp. 24--es.

\bibitem{wessel2018tdemaccuracy}
B.~Wessel, M.~Huber, C.~Wohlfart, U.~Marschalk, D.~Kosmann, and A.~Roth, ``Accuracy assessment of the global {TanDEM-X} digital elevation model with {GPS} data,'' \emph{ISPRS Journal of Photogrammetry and Remote Sensing}, vol. 139, pp. 171--182, 2018.

\bibitem{colmap1}
J.~L. Sch\"{o}nberger and J.-M. Frahm, ``{Structure-from-Motion Revisited},'' in \emph{Conference on Computer Vision and Pattern Recognition (CVPR)}, 2016.

\bibitem{colmap2}
J.~L. Sch\"{o}nberger, E.~Zheng, M.~Pollefeys, and J.-M. Frahm, ``{Pixelwise View Selection for Unstructured Multi-View Stereo},'' in \emph{European Conference on Computer Vision (ECCV)}, 2016.

\bibitem{guizilini2022learning}
V.~Guizilini, K.-H. Lee, R.~Ambru{\c{s}}, and A.~Gaidon, ``Learning optical flow, depth, and scene flow without real-world labels,'' \emph{IEEE Robotics and Automation Letters}, vol.~7, no.~2, pp. 3491--3498, 2022.

\bibitem{teed2020raft}
Z.~Teed and J.~Deng, ``Raft: Recurrent all-pairs field transforms for optical flow,'' in \emph{Computer Vision--ECCV 2020: 16th European Conference, Glasgow, UK, August 23--28, 2020, Proceedings, Part II 16}.\hskip 1em plus 0.5em minus 0.4em\relax Springer, 2020, pp. 402--419.

\bibitem{shin2023ir}
U.~Shin, K.~Park, B.-U. Lee, K.~Lee, and I.~S. Kweon, ``Self-supervised monocular depth estimation from thermal images via adversarial multi-spectral adaptation,'' in \emph{Proceedings of the IEEE/CVF Winter Conference on Applications of Computer Vision}, 2023, pp. 5798--5807.

\bibitem{zhang2021deep}
R.~Zhang, L.~Xu, Z.~Yu, Y.~Shi, C.~Mu, and M.~Xu, ``Deep-irtarget: An automatic target detector in infrared imagery using dual-domain feature extraction and allocation,'' \emph{IEEE Transactions on Multimedia}, vol.~24, pp. 1735--1749, 2021.

\bibitem{zhang2024part}
R.~Zhang, J.~Tan, Z.~Cao, L.~Xu, Y.~Liu, L.~Si, and F.~Sun, ``Part-aware correlation networks for few-shot learning,'' \emph{IEEE Transactions on Multimedia}, 2024.

\bibitem{zhang2024cognition}
R.~Zhang, Z.~Cao, S.~Yang, L.~Si, H.~Sun, L.~Xu, and F.~Sun, ``Cognition-driven structural prior for instance-dependent label transition matrix estimation,'' \emph{IEEE Transactions on Neural Networks and Learning Systems}, pp. 1--14, 2024.

\end{thebibliography}

\vfill\eject

\begin{IEEEbiography}
[{\includegraphics[width=1in,height=1.25in,clip,keepaspectratio]{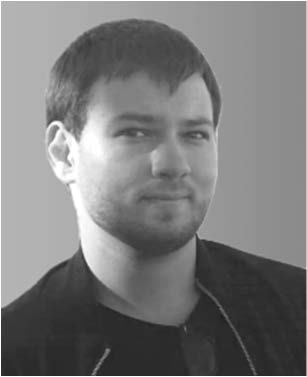}}]
{Horatiu Florea} received the engineer and M.S. degrees in the field of Computer Science from the Technical University of Cluj-Napoca in 2017 and 2019, respectively. He is currently pursuing a Ph.D. degree in Computer Science at the same university, specializing in Computer Vision approaches for 3D environment perception. His research interests include data acquisition and synchronization, LiDAR processing, 3D reconstruction and monocular depth estimation with applications in Automated Driving and Unmanned Aerial Systems.
\end{IEEEbiography}

\begin{IEEEbiography}
[{\includegraphics[width=1in,height=1.25in,clip,keepaspectratio]{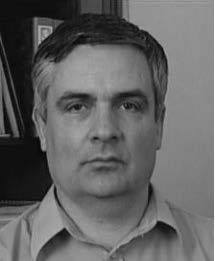}}]
{Sergiu Nedevschi} (Member, IEEE) received the M.S. and Ph.D. degrees in electrical engineering from the Technical University of Cluj-Napoca (TUCN),
Romania, in 1975 and 1993, respectively. From 1976 to 1983, he was a Researcher with the Research Institute for Computer Technologies Cluj-Napoca. Since 1983, he has been with TUCN. He was appointed as a Full Professor of computer science, in 1998, he founded and  since 1998, he has been leading the Image Processing and Pattern Recognition Research Center. From 2000 to 2004, he was the Head of the Computer Science Department. From 2004 to 2012, he was the Dean of the Faculty of Automation and Computer Science. From 2012 to 2020, he was the Vice-Rector with Scientific Research of TUCN. He was involved in more than 80 research projects, being the coordinator of 62 of them. His industrial cooperation with important automotive players, such as Volkswagen AG, Robert Bosch GmbH, SICK AG, and research institutes, such as VTT, INRIA was achieved through funded research projects. He has published more than 400 scientific articles and has edited over 20 volumes, including books and conference proceedings. His research interests include image processing, pattern recognition, computer vision, machine learning, intelligent, and autonomous vehicles.
\end{IEEEbiography}

\end{document}